\documentclass[journal]{IEEEtran}
\usepackage{amssymb}
\usepackage{multirow}
\usepackage{times}
\usepackage{helvet}
\usepackage{courier}
\usepackage[hyphens]{url}
\usepackage{graphicx}
\usepackage{booktabs}
\usepackage{amsmath,bm}
\usepackage {mathtools}
\usepackage{threeparttable}
\usepackage{marvosym}
\usepackage{makecell}
\usepackage{color}

\ifCLASSINFOpdf
\else
\fi
\begin{document}
%
% paper title
% Titles are generally capitalized except for words such as a, an, and, as,
% at, but, by, for, in, nor, of, on, or, the, to and up, which are usually
% not capitalized unless they are the first or last word of the title.
% Linebreaks \\ can be used within to get better formatting as desired.
% Do not put math or special symbols in the title.
\title{Actor and Action Modular Network for Text-based Video Segmentation}
%
%
% author names and IEEE memberships
% note positions of commas and nonbreaking spaces ( ~ ) LaTeX will not break
% a structure at a ~ so this keeps an author's name from being broken across
% two lines.
% use \thanks{} to gain access to the first footnote area
% a separate \thanks must be used for each paragraph as LaTeX2e's \thanks
% was not built to handle multiple paragraphs
%

\author{Jianhua~Yang, Yan~Huang, Kai Niu, Linjiang Huang,
        Zhanyu~Ma,~\IEEEmembership{Senior~Member, ~IEEE},
        and~Liang~Wang\textsuperscript{\Letter},~\IEEEmembership{~Fellow,~IEEE}% <-this % stops a space

\thanks{J. Yang and Z. Ma are with the Pattern Recognition and Intelligent Systems Lab., School of Artificial Intelligence, Beijing University of Posts and Telecommunications, Beijing, China (E-mail: \{youngjianhua, mazhanyu \}@bupt.edu.cn); K. Niu is with the School of Computer Science, Northwestern Polytechnical University, Xi'an, China, (E-mail: kai.niu@nwpu.edu.cn); L. Huang is with the Multimedia Laboratory, Chinese University of Hong Kong, (E-mail: ljhuang524@gmail.com); Y. Huang and L. Wang are with the Center for Research on Intelligent
Perception and Computing (CRIPAC), National Laboratory of Pattern Recognition (NLPR), Institute of Automation, Chinese Academy of Sciences (CASIA), Beijing, China, and with the University of Chinese Academy of Sciences (UCAS), Beijing, China. (E-mail: \{yhuang, wangliang\}@nlpr.ia.ac.cn)}
\thanks{Corresponding author: Liang Wang}
}

% The paper headers
\markboth{Journal of \LaTeX\ Class Files,~Vol.~14, No.~8, August~2015}%
{Shell \MakeLowercase{\textit{et al.}}: Bare Demo of IEEEtran.cls for IEEE Journals}
% The only time the second header will appear is for the odd numbered pages
% after the title page when using the twoside option.
%
% *** Note that you probably will NOT want to include the author's ***
% *** name in the headers of peer review papers.                   ***
% You can use \ifCLASSOPTIONpeerreview for conditional compilation here if
% you desire.

% If you want to put a publisher's ID mark on the page you can do it like
% this:
%\IEEEpubid{0000--0000/00\$00.00~\copyright~2015 IEEE}
% Remember, if you use this you must call \IEEEpubidadjcol in the second
% column for its text to clear the IEEEpubid mark.

% use for special paper notices
%\IEEEspecialpapernotice{(Invited Paper)}

% make the title area
\maketitle

% As a general rule, do not put math, special symbols or citations
% in the abstract or keywords.
\begin{abstract}
Text-based video segmentation aims to segment an actor in video sequences by specifying the actor and its performing action with a textual query. Previous methods fail to explicitly align the video content with the textual query in a fine-grained manner according to the actor and its action, due to the problem of \emph{semantic asymmetry}. The \emph{semantic asymmetry} implies that two modalities contain different amounts of semantic information during the multi-modal fusion process. To alleviate this problem, we propose a novel actor and action modular network that individually localizes the actor and its action in two separate modules. Specifically, we first learn the actor-/action-related content from the video and textual query, and then match them in a symmetrical manner to localize the target tube. The target tube contains the desired actor and action which is then fed into a fully convolutional network to predict segmentation masks of the actor. Our method also establishes the association of objects cross multiple frames with the proposed temporal proposal aggregation mechanism. This enables our method to segment the video effectively and keep the temporal consistency of predictions. The whole model is allowed for joint learning of the actor-action matching and segmentation, as well as achieves the state-of-the-art performance for both single-frame segmentation and full video segmentation on A2D Sentences and J-HMDB Sentences datasets.

% Extensive experiments on two benchmark datasets, Actor-Action Dataset Sentences (A2D Sentences) and J-HMDB Sentences, demonstrate that our proposed approach outperforms state-of-the-art methods.
%Extensive experimental results on both the actor-acion dataset and the Youtube-objects datasets demonstrate that the proposed approach outperforms the state-of-the-art methods.
% In this paper, we explore the problem of referring expression comprehension from the perspective of language-driven visual reasoning, and propose a ...model
% which poses great potential for various applications such as video surveillance.

\end{abstract}

% Note that keywords are not normally used for peerreview papers.

% For peer review papers, you can put extra information on the cover
% page as needed:
% \ifCLASSOPTIONpeerreview
% \begin{center} \bfseries EDICS Category: 3-BBND \end{center}
% \fi
%
% For peerreview papers, this IEEEtran command inserts a page break and
% creates the second title. It will be ignored for other modes.
\IEEEpeerreviewmaketitle

\section{Introduction}
 With the explosion of video data on the Internet, understanding video content becomes increasingly important and has attracted significant research interest in recent years. Action recognition is one of the key tasks in the field of video analysis, which mainly focuses on human-centric action classification \cite{niebles2007hierarchical,girdhar2017actionvlad} or localization \cite{klaser2010human,shou2016temporal} in videos. Since traditional action recognition lacks fine-grained understanding of video content, recently there is a growing interest in simultaneously understanding actors and actions in videos. Xu \emph{et al.} \cite{xu2015can} first proposed a new actor-action segmentation challenge on a large-scale video dataset, \emph{i.e.,} actor-action dataset (A2D). Actor-action semantic segmentation requires to spatio-temporally localize and recognize seven classes of actors (\emph{e.g.}, ``adult'' and ``bird'') and eight classes of actions (\emph{e.g.}, ``climb'' and ``fly'') at pixel-level in a video. Various approaches under supervision \cite{xu2016actor,qiu2017learning,kalogeiton2017joint,dang2018actor,ji2018end,Rana_2021_WACV} or weak supervision \cite{yan2017weakly,Chen2020Learning} have been proposed to tackle this problem and achieved significant advance. However, the actors and actions of interest are much more diverse in the real world, thus the small number of pre-defined actor and action classes largely limits applications of the aforementioned methods in some cases, such as automatic video editing, intelligent vision search, and human-robot interaction.

\begin{figure}
\centering
\includegraphics[width=8.6cm]{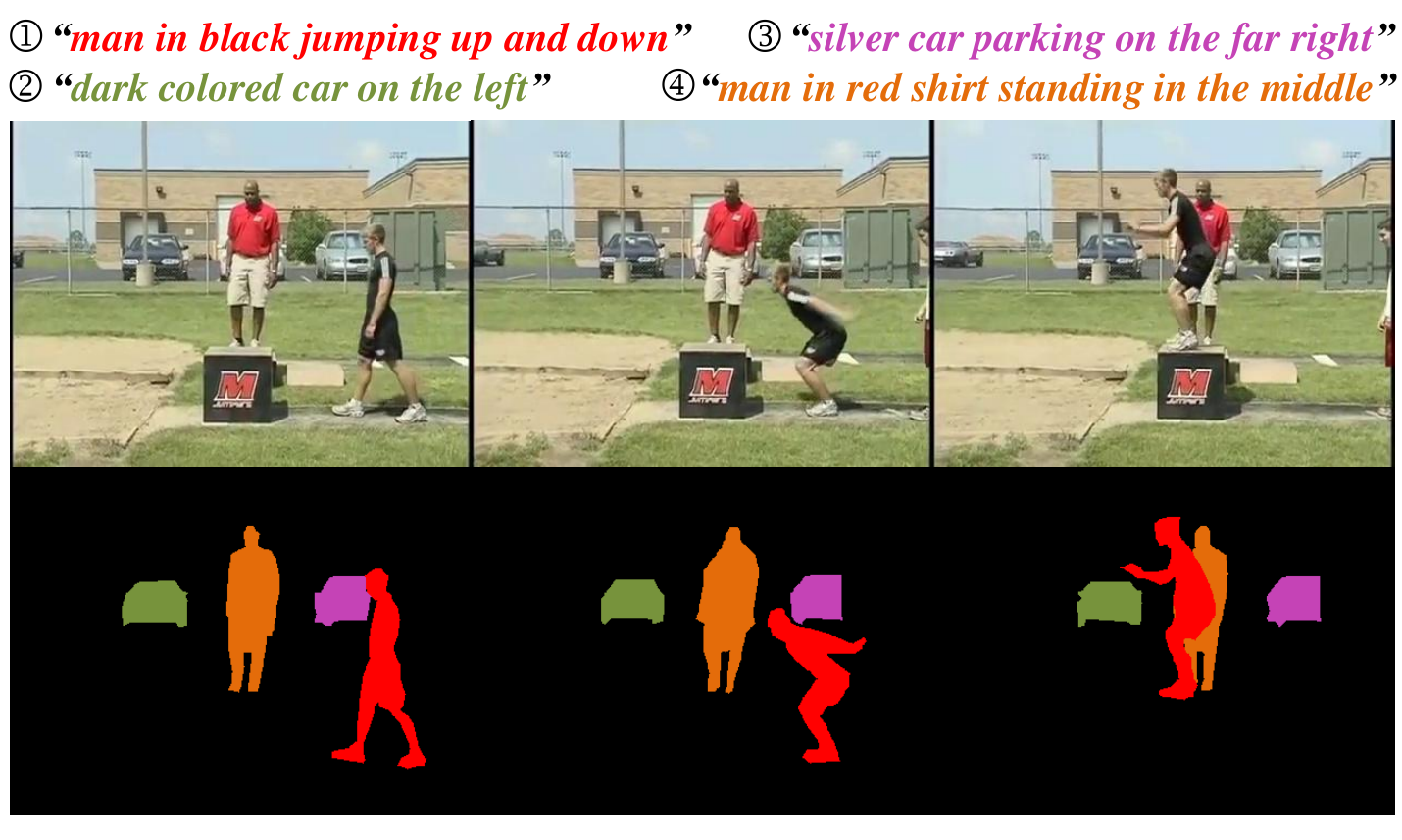}
\caption{Illustration of text-based video segmentation. The goal of the task is to attain segmentation masks of the actor in video sequences by specifying the actor and its action with a textual query. Note that the mask color corresponds to the textual query color (Better viewed in color).}
%Given a video and corresponding textual queries which specify the actors and they are performing actions, the goal of this task is to produce binary masks of query-specified actors in video frames. Note that the mask color corresponds to the textual query color (Better viewed in color).}
\label{fig:gifure1}
\vspace{-0.2cm}
\end{figure}

\begin{figure*}[t]
\centering
\includegraphics[width=16.0cm]{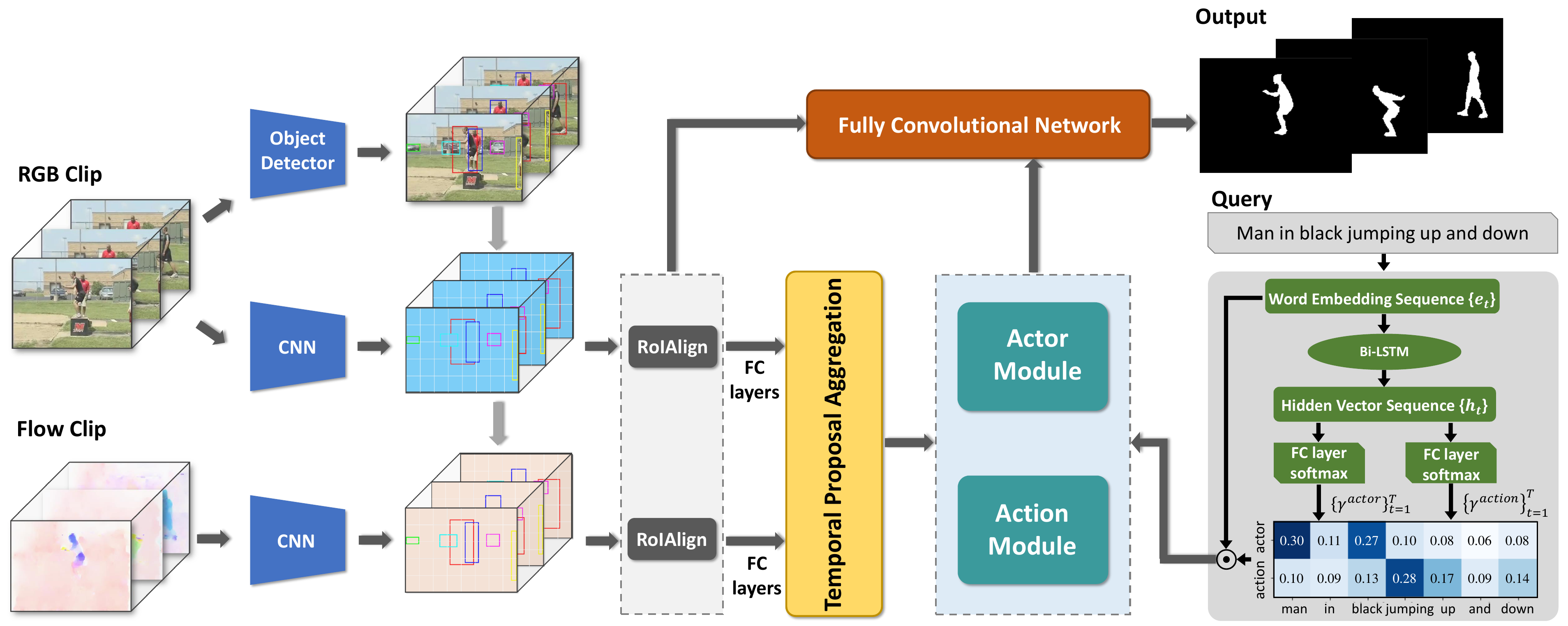}  % 16.0cm
\vspace{-0.2cm}
\caption{An overview of our proposed actor and action modular network (AAMN) for text-based video segmentation (Better viewed in color). The AAMN takes a textual query, RGB and Flow clips as well as corresponding proposals generated from an external detector as inputs. The temporal proposal aggregation mechanism will generate a set of actor-/action-related tubes by computing linking scores. The language attention model will adaptively learn the actor-/action-related representation from the given textual query. Two modules are proposed to localize the target tube that involves the actor and action referred to by the textual query. The fully convolutional network (FCN) is used to predict the masks of the actor within the target tube.}
\label{fig:framework}
\vspace{-0.4cm}
\end{figure*}

Compared with word-level actor and action classes, textual queries are more flexible so that they can be used to specify various actors and actions of interest. While most related works focus on either spatial object localization from textual queries in static images \cite{yu2016modeling,hu2016segmentation,liu2017recurrent} or temporal action localization from textual queries in videos \cite{anne2017localizing,gao2017tall}. There are few works that consider the actor and the corresponding action simultaneously in a video when the actor and action are specified by a given textual query. To bridge this gap, the new task of text-based video segmentation was originally introduced by Gavrilyuk \emph{et al.} \cite{gavrilyuk2018actor},  whose goal was to attain pixel-level segmentation of the actor in video content by specifying the actor and its performing action with a textual query. Some illustrative examples of the new task are shown in Fig. \ref{fig:gifure1}, where four textual queries are given to specify four different actors and their actions, respectively. The model is required to predict segmentation masks of actors frame-by-frame conditioned by the textual queries. The variety of textual queries provides a flexible way to select the actors of interest according to actors' appearance information and motion information. Appearance information contains visual cues about categories, colors, and shapes of actors, while motion information represents actions that are performed by actors in the video. Considering either video or textual query understanding has been extensively studied in its related fields, the major challenge of this task is to establish a suitable video-query alignment by associating with actors and actions. Despite much progress has been achieved in this task, existing methods \cite{gavrilyuk2018actor,wang2019asymmetric,wang2020context,mcintosh2018multi} suffer from the following two aspects.

First, the fine-grained alignment between video and query features is not explored well. Existing methods adopt a bottom-up paradigm to address the text-based video segmentation, they fail to capture entity-level (\emph{i.e.}, actor and action) information for explicitly modeling the fine-grained alignment between two modalities. More concretely, these methods first integrate video and query features with various fusion strategies, and then they implicitly align each convolved local volume (\emph{i.e.}, Spatio-temporal feature unit) of the 3D feature map produced by a 3D convolutional neural network (CNN) with a holistic representation of the textual query. Finally, they measure the binary similarity map from the decoder network. However, for the query representation, the arbitrary input query could include various words referring to actors, actions, attributes, locations, objects, \emph{etc.}, which contains much actor-/action-related information. Besides, the variation of linguistic structures of queries makes these methods difficult to precisely capture the semantic information related to the actor and its action. For the video features, the convolved local volume usually provides either redundant or incomplete visual content for a single actor depending on relative sizes of the actor and convolutional filters. In addition, since each local volume could not include sufficient information to recognize an action, modeling spatio-temporal relationships among local volumes \cite{he2019stnet, nonlocal2018} is helpful to recognize the action. Therefore, the local volume of the video features and the holistic representation of the textual query contain different amounts of semantic information, namely, \emph{semantic asymmetry} between the local volume and the query representation. The \emph{semantic asymmetry} fails to align the textual query with referred actor and its action in video content, and eventually degrades segmentation results.

Second, existing methods are inefficient for full video segmentation and suffer from the discrepancy of predictions in consecutive frames, as they do not consider object association between adjacent frames. As A2D dataset \cite{xu2015can} is sparsely annotated 3 to 5 frames for a video, existing methods \cite{gavrilyuk2018actor,wang2019asymmetric,wang2020context} directly aggregated the spatio-temporal features into the annotated frames to train and evaluate their models. Consequently, there are two limitations for these methods: 1) They are time-consuming to segment all frames of the video by densely sampling lots of video clips, because their models only predict one-frame segmentation results using 2D decoder networks with multi-frame video clips as inputs. 2) The temporal consistency of predictions cross multiple frames is not established well as the outputs of the models are treated independently. The scene changes and motion blur would lead to the discrepancy of predictions in consecutive frames. To alleviate these issues, McIntosh \emph{et al.} \cite{mcintosh2018multi} replaced 2D deconvolutional layers with 3D deconvolutional layers in the decoder network, and used annotations of bounding boxes to train and evaluate their model. However, the block-like outputs from the model are not precise enough for pixel-level actor segmentation in videos.

To overcome the limitation of \emph{semantic asymmetry}, an intuitive idea is to first find actor-/action-related video content and words, rather than equally split local volumes of video features and directly use the holistic representation from the textual query. This can be easily achieved by leveraging the recent advances of object detection \cite{liu2016ssd,ren2015faster,he2017mask} and language attention mechanism \cite{hu2017modeling,yu2018mattnet,wang2019neighbourhood}. Then the cross-modal matching between extracted instance-level regions and actor-related words is straightforward, as it is similar to the referring expression methods \cite{yu2018mattnet,wang2019neighbourhood}. If the region contains the actor referred to by the query, namely, \emph{semantic symmetry} between the region and the actor-related words, it is desired region. However, the referring expression methods have two drawbacks for text-based video segmentation. On the one hand, the referring expression methods do not refer to the dynamic information or actions of the actor, the action-related matching is still a challenge. The action-related matching additionally requires understanding the dynamic motion rather than only static actors. It is more complicated as the video content also includes other disturbing relationships, \emph{e.g.}, affiliation and position, which can easily lead to confusion. Moreover, the actor-related matching and action-related matching are closely related, \emph{i.e.}, it is the actor that actually performs the action. But how to make them collaborate with each other to improve the final performance is rarely investigated. On the other hand, the referring expression methods fail to establish data association for objects cross multiple frames. These methods process the video frame-by-frame independently without considering the valuable temporal information cross frames, which could result in the discrepancy of predictions due to the scene or appearance variations in the video.

In this paper, we propose a novel model which named Actor and Action Modular Network (AAMN) for text-based video segmentation. Specifically, to effectively utilize the limited annotations and establish the association of objects in consecutive frames, we first generate a set of actor-/action-related tubes by introducing a temporal proposal aggregation mechanism. This allows us to obtain appearance and motion representations for each tube of the clip. In order to realize semantically symmetrical matching, we implement the language attention mechanism on the textual query to adaptively focus on actor-/action-related words, and then construct two modules in AAMN. One module is called actor module, which is designed to identify the actor based on actor's appearance information. It associates actor-related tubes with the corresponding words to produce actor matching scores. Another module is called action module, which is designed to identify the actor based on actions performed by the actor in the video. This module utilizes optical flow features to model the dynamic motion information of the actor. To further keep our model from being disturbed by ambiguous motion information or background, the action module contains a contextual long short-term memory network (LSTM) to model the potential relationships among pairwise tubes. The action module integrates motion features and contextual features as well as action-related words to predict the action-matching scores. Finally, the feature maps of the target tube are fed into a small fully convolutional network (FCN) \cite{long2015fully} to predict masks of the input clip. Our method can not only achieve single-frame video segmentation like previous methods in \cite{gavrilyuk2018actor,wang2019asymmetric,wang2020context}, but also full video segmentation like the method in \cite{mcintosh2018multi}. To demonstrate the effectiveness of the proposed AAMN, we conduct experiments on two benchmark datasets, and experimental results show that our method outperforms the state-of-the-art methods.

The contributions of this paper are highlighted as follows:
\begin{itemize}
    \item [i)]
    We study on a practically important problem, \emph{i.e.}, text-based video segmentation, and accordingly propose a novel Actor and Action Modular Network (AAMN) to deal with the intrinsic \emph{semantic asymmetric} problem.
    \item [ii)]
    We propose a temporal proposal aggregation mechanism to establish objects association cross adjacent frames. This enables AAMN to achieve multi-frame segmentation at a time with limited annotations, and keep the temporal consistency of predictions in consecutive frames.
    \item [iii)]
    We especially focus on the action-related matching between the video content and the textual query, by constructing an action module to jointly model the dynamic motion and visual context.
    \item [iv)]
    We quantitatively and qualitatively validate the effectiveness of our method and achieve the state-of-the-art results on A2D Sentences and J-HMDB Sentences datasets.
\end{itemize}

\section{Related Work}
\subsection{Actor-Action Semantic Segmentation in Videos}

There are many emerging works on video object segmentation that aims to segment out a particular object in an entire video \cite{yao2020video}. Actor-action semantic segmentation in videos differs from them by assigning an action label to the target actor. Such a detailed actor-action understanding task has attracted growing attention in recent years. Xu \emph{et al.} made the first effort on actor-action semantic segmentation problem in \cite{xu2015can}, where they collected a large-scale video dataset, \emph{i.e.,} A2D, to jointly consider various types of actors performing various actions. Early works \cite{xu2015can,xu2016actor} solved this problem based on Conditional Random Fields (CRF) with supervoxels features. With the success of deep learning in the field of computer vision, deep learning-based approaches have been widely studied for actor-action semantic segmentation. Qiu \emph{et al.} \cite{qiu2017learning} exploited 2D and 3D FCN to model spatio-temporal dependency from a video for semantic segmentation. Dang \emph{et al.} \cite{dang2018actor} relied on region masks to enhance consistent labeling of action for actor-action semantic segmentation. Kalogeiton \emph{et al.} \cite{kalogeiton2017joint} proposed a multitask architecture for joint actor-action detection and then performed actor-action segmentation with SharpMask \cite{pinheiro2016learning}. Ji \emph{et al.} \cite{ji2018end} leveraged multiple input modalities and contextual information from videos to realize pixel-wise actor-action segmentation and detection in a unified network with joint multitask learning. Rana \emph{et al.} \cite{Rana_2021_WACV} proposed an effective method to perform pixel-level actor-action detection in a single-shot. Moreover, the weakly supervised actor-action semantic segmentation, which only has access to video-level actor and action tags, has also been studied in recent works \cite{yan2017weakly,Chen2020Learning}.

Although promising results have been achieved by the aforementioned approaches, the pre-defined actor and action categories limit their applications in human-computer interaction scenarios. Our work focuses on solving a more challenging task, namely, text-based video segmentation, which aims to segment a particular actor specified by a textual query referring to the actor and its performing action in the video.

\subsection{Referring Expression Comprehension and Segmentation}

The referring expression comprehension (REC) aims to localize the object with a bounding box in the image that corresponds to a given textual query \cite{mao2016generation}. To address this problem, most of proposed approaches focus on mining contextual information from the language and image \cite{yu2016modeling, li2018bundled}, modeling the relationship between different objects \cite{wang2019neighbourhood, yang2019dynamic}, and analyzing the linguistic structures \cite{yu2018mattnet}. To precisely describe the shape of the referred object, the referring expression segmentation (RES) goes a step further to predict the segmentation mask instead of the bounding box. Hu \emph{et al.} first proposed the baseline by up-sampling the concatenation of visual and linguistic features using a deconvolutional layer. Margffoy-Tuay \emph{et al.} \cite{margffoy2018dynamic} integrated the visual and linguistic features by generating the dynamic filter for each word of the textual query. Li \emph{et al.} \cite{li2018referring} explored to fuse multi-level visual features to recurrently refine the local details of segmentation masks. To enable better multi-modal interactions, more recent methods \cite{shi2018key-word-aware,chen2019see-through-text,cmsa_pami,hu2020bi,liu2021cross,yang2021cmf} focus on establish complex cross-modal relations between visual and linguistic features. For example, Ye \emph{et al.} \cite{cmsa_pami} integrated every word with multi-level visual features and introduce self-attention to adaptively focus on informative words of query and regions of the image. Hu \emph{et al.} \cite{hu2020bi} proposed a bi-directional attention module to establish cross-modal correlations, Liu \emph{et al.} \cite{liu2021cross} leveraged attention mechanism to capture different concepts for enhancing the multi-modal fusion process.

Text-based video segmentation is closely related to the task of RES, some works \cite{cmsa_pami,liu2021cross} also explored to address text-based video segmentation with RES methods. However, there are still significant differences between these two tasks. \emph{First}, for the RES, the target object is mainly referred to by its appearance or location in a given static image. While for text-based video segmentation, the target object is referred to by not only its attributes (\emph{e.g.}, colors, shapes), but also its performing actions in the video. This indicates that the text-based video segmentation is more complicated than RES. \emph{Second}, RES methods segment the video frame-by-frame without considering the data association between adjacent frames, which usually leads to the discrepancy of predicted objects. It is essential to establish the association of objects cross multiple frames for text-based video segmentation. \emph{Third}, text-based video segmentation adopts more metrics to evaluate the segmentation performance except the overall intersection over union that is often used in RES.

\subsection{Text-based Video Segmentation}

The task of text-based video segmentation was originally introduced by Gavrilyuk \emph{et al.} \cite{gavrilyuk2018actor}, whose goal was to segment the actor specified by a textual query referring to the actor and its performing action in the video. Existing approaches address this problem by following a bottom-up procedure, which extracts the video and query features separately and then predicts the segmentation mask from the merged heterogeneous features. Specifically, Gavrilyuk \emph{et al.} \cite{gavrilyuk2018actor} proposed to correlate the video contents and sentences using dynamic convolutions. Wang \emph{et al.}  \cite{wang2019asymmetric} introduced vision guided language attention and language guided vision attention mechanisms to obtain a robust language representation and aggregate global visual context for performance gains. Wang \emph{et al.} \cite{wang2020context} constructed a context modulated dynamic convolutional network to incorporate contextual information into the correlation learning of heterogeneous modalities. McIntosh \emph{et al.} \cite{mcintosh2018multi} further encoded the visual and language features as capsules and integrated the visual and language information via a routing algorithm. Unlike aforementioned works, we propose to solve the text-based video segmentation from another perspective, namely, a top-down procedure. We first localize the referred actor and action with bounding boxes sequence in a symmetric matching way, then predict the masks of the actor along the bounding boxes sequence. Our method can effectively localize the query described actor and action, and reduce disturbances of irrelevant objects and the background in videos.

\section{Methodology}
In this section, we elaborate on the proposed method for text-based video segmentation, the overall framework of our method is illustrated in Fig. \ref{fig:framework}.

\textbf{Task definition.} Given a video sequence $\mathcal{I}=\{I_{n}\in{\mathbb{R}^{W\times{H}\times{3}}}\}_{n=1}^{N}$ with $N$ frames, and a corresponding textual query $\mathcal{W}=\{w_{t}\}_{t=1}^{T}$ with $T$ words. The textual query usually refers to the actor and its performing action in the video. The text-based video segmentation aims to produce $N$-frame binary segmentation masks $\mathcal{M}=\{M_{n}\}_{n=1}^{N}$, $M_{n} \in{\mathbb{R}^{H\times{W}}}$ for the referred actor in the video sequence.

\textbf{Overview.}
Except appearance information, which contains visual cues about categories, colors and shapes of the actors, dynamic motion information is another important visual cue to identify the actors in the video. This motivated us to localize the actor cross frames by aligning appearance information and motion information between the video content and the textual query. To explicitly model such fine-grained alignment and achieve semantically symmetric matching between two modalities, we propose an actor and action modular network (AAMN) for text-based video segmentation.

Formally, our AAMN first generates a set of proposals per frame in a video clip via an external object detector. Then AAMN feeds RGB and Flow clips as well as corresponding proposals into two parallel CNN architectures to learn the appearance and motion representations for each proposal. To link the proposals cross frames and generate a set of actor-/action-related tubes for input clips, a temporal proposal aggregation mechanism is introduced after fully connected (FC) layers (Sec. \ref{section: tube_generation}). Meanwhile, in order to obtain related representations about the actor and action from the textual query, AAMN decomposes the query into two separate components relying on a language attention mechanism (Sec. \ref{section: attn_learn}). Afterward, we construct two modules, \emph{i.e.}, actor module and action module, to identify the actor based actor's appearance information and motion information, respectively. These two modules take actor-/action-related tubes and language representations as inputs and localize the target tube according to the actor and action matching scores (Sec. \ref{section: modular_network}). Finally, a tiny FCN is followed by two modules to predict the segmentation masks of the actor within the target tube (Sec. \ref{section: segmentation_network}). We train the whole network with a multi-task learning strategy by combining the actor-action matching and the actor segmentation (Sec. \ref{section: training_model}).

\subsection{Actor-/Action-related Tubes Generation and Representation}
\label{section: tube_generation}

The AAMN takes RGB and Flow clips as well as pre-generated proposals as inputs, where each clip contains 2\emph{L}+1 frames. For proposals generation, we follow the previous work \cite{kalogeiton2017joint} to choose the commonly used object detector, \emph{i.e.}, Faster R-CNN \cite{ren2015faster}, to extract a fixed number of proposals with high confidence in each frame of the video. We simply denote these proposals as $\{\{r_k^{n}\}_{k=1}^{K}\}_{n=1}^{N}$, where the video has \emph{N} frames and each frame contains \emph{K} proposals. $r_{k}^{n}$ denotes the \emph{k}-th proposal in the frame \emph{n}, and it is comprised of the top-left and bottom-right coordinates. The Faster R-CNN model has been trained on a large-scale detection dataset MSCOCO \cite{lin2014microsoft}. In our experiments, we further finetune the detector on a new dataset, \emph{i.e.}, A2D, to improve the detection accuracy for actors.

Considering that only 3 to 5 frames are annotated in videos of A2D dataset, we sample 3 to 5 clips around the annotated frames from a video to train our model. For each clip, the middle frame, \emph{i.e.}, the (\emph{L}+1)-th frame is annotated with bounding box and binary mask, while forward \emph{L} frames and backward \emph{L} frames are not annotated. To localize and segment the query-specified actor in consecutive frames of the clip, we first link the detected proposals $\{\{r_{k}^{n}\}_{k=1}^{K}\}_{n=1}^{2L+1}$ in temporal dimension and generate \emph{K} tubes (\emph{i.e.}, bounding boxes sequences). To this end, we introduce a temporal proposal aggregation mechanism that expects two proposals in adjacent frames contain the same actor or object if they possess similar spatial locations and features. Specifically, we utilize a CNN architecture to extract the appearance features in RGB stream and further obtain fixed-size (\emph{e.g., $7\times{7}$}) feature maps via RoIAlign operation \cite{he2017mask} for the generated proposals. Similar to Mask R-CNN \cite{he2017mask} which has multi-task heads for classes prediction, bounding boxes regression, and object segmentation, the small feature maps in RGB stream are fed into two different branches for actor-action matching and segmentation, respectively. For actor-action matching branch, we transform each small map into a holistic appearance representation $\bm{v}_{k}^{n}$ $\in{\mathbb{R}^{C_{v}}}$ ($k\in{[1,K]}, n\in{[1,2L+1]}$) for the \emph{k}-th proposal in the \emph{n}-th frame via two FC layers. Similarly, we can obtain the holistic motion representation $\bm{f}_{k}^{n}$ $\in{\mathbb{R}^{C_{v}}}$ ($k\in{[1,K]}, n\in{[1,2L+1]}$) for the \emph{k}-th proposal in the \emph{n}-th frame in Flow stream.

Next, the linking score between proposals $r_{k_{1}}^{n_{1}}$ and $r_{k_{2}}^{n_{2}}$ is formulated as:
\begin{equation}
\label{equ:linking_score}
s_{link}(r_{k_{1}}^{n_{1}}, r_{k_{2}}^{n_{2}}) = IoU(r_{k_{1}}^{n_1}, r_{k_{2}}^{n_2}) + \dfrac{\rho}{1+ED(\bm{v}_{k_{1}}^{n_{1}}, \bm{v}_{k_{2}}^{n_{2}})},
\end{equation}
where the $IoU(\cdot)$ is utilized to measure the spatial similarity by computing the intersection-over-union between proposals $r_{k_{1}}^{n1}$ and $r_{k_{2}}^{n2}$. The features of $r_{k_{1}}^{n1}$ and $r_{k_{2}}^{n2}$ are denoted as $\bm{v}_{k_{1}}^{n_{1}}$ and $\bm{v}_{k_{2}}^{n_{2}}$, respectively. The $ED(\cdot)$ means the Euclidean distance which is adopted to measure the semantic similarity between features $\bm{v}_{k_{1}}^{n_{1}}$ and $\bm{v}_{k_{2}}^{n_{2}}$. The $\rho$ is a balanced parameter. For each proposal in the $n_{1}$-th frame, we select the proposal with maximal linking score from the $n_{2}$-th frame. In this way, we can obtain \emph{K} tubes for each video clip. Note that we only use visual features from RGB stream to calculate the semantic similarity.
%The main reason is that the appearance information is consistent within a video clip, while visual features from optical flow cannot keep the motion information in each frame of the clip.

To obtain the representation of each tube in the RGB stream, we apply mean pooling over features from forward \emph{L} frames and backward \emph{L} frames to aggregate them as temporal context of the (\emph{L}+1)-th frame, and perform addition with the feature of the (\emph{L}+1)-th frame:
\begin{equation}
\label{equ:avg_pool}
\widetilde{\bm{v}}_{k}^{L+1} = \bm{v}_{k}^{L+1} + \dfrac{1}{2L}\sum_{n=1; n\neq{L+1}}^{2L+1} \bm{v}_{k}^{n}.
\end{equation}

Finally, we use one FC layer with Rectified Linear Unit (ReLU) function to transform $\widetilde{\bm{v}}_{k}^{L+1}$ and obtain the appearance representation for the \emph{k}-th tube. For simplicity, we denote the representation as $\bm{v}_{k}$ $\in{\mathbb{R}^{C_{v}^{'}}}$. In Flow stream, we apply the same way to obtain the motion representation $\bm{f}_{k}$ $\in{\mathbb{R}^{C_{v}^{'}}}$ for the \emph{k}-th tube in the Flow clip.

\subsection{Textual Query Representation Learning}
\label{section: attn_learn}

As AASTQ takes an arbitrary textual query as input, the query may contain diversely linguistic structures or irrelevant semantics with relation to the actor and its action. To alleviate the interference of the linguistic structures and irrelevant semantics, a straightforward idea is to utilize external language parser \cite{socher2013parsing,andreas2016learning} or pre-defined templates \cite{kazemzadeh2014referitgame} to parse the query into different components and then extract the actor-/action-related individual words for the actor and action localization. Nevertheless, previous works \cite{hu2017modeling,yu2018mattnet,2020graph} have demonstrated that the language parser cannot work well in the query-region matching problem, because individual words lack essential contextual information for scene understanding of images. As an alternative way, Hu \emph{et al.} \cite{hu2017modeling} proposed to automatically parse the query into a triplet (\emph{subject}, \emph{relationship}, \emph{object}) with a soft attention mechanism for relational reasoning. Similar to \cite{hu2017modeling}, Yu \emph{et al.} \cite{yu2018mattnet} learned to parse the query into a triplet (\emph{subject}, \emph{location}, \emph{relationship}) for object localization. In our work, we aim to explicitly model the actor and action localization from a given query. Hence, we adopt the attention mechanism to adaptively learn to attend to the words that are relevant to the actor and action, respectively.

More specifically, for a given textual query with $T$ words $\{w_{t}\}_{t=1}^T$, we use Glove \cite{pennington2014glove} to embed each word and obtain a vector sequence $\{\bm{e}_{t}\}_{t=1}^T$ $\in{\mathbb{R}^{C_{e}}}$. Then a two-layer Bi-directional LSTM \cite{schuster1997bidirectional} takes the vector sequence $\{\bm{e}_{t}\}_{t=1}^T$ as input and encodes the query as follows:

\begin{equation}
\begin{split}
&\overrightarrow{\bm{h}_{t}^{1}}=\overrightarrow{\rm{LSTM}}\left(\overrightarrow{\bm{h}_{t-1}^{1}}, \ \bm{e}_{t}\right);  \overleftarrow{\bm{h}_{t}^{1}}=\overleftarrow{\rm{LSTM}}\left(\overleftarrow{\bm{h}_{t+1}^{1}}, \ \bm{e}_{t}\right), \\
&\bm{h}_{t}^{1}=\left[ \overrightarrow{\bm{h}_{t}^{1}}, \ \overleftarrow{\bm{h}_{t}^{1}} \right], \\
&\overrightarrow{\bm{h}_{t}^{2}}=\overrightarrow{\rm{LSTM}}\left(\overrightarrow{\bm{h}_{t-1}^{2}}, \ \bm{h}_{t}^{1}\right);  \overleftarrow{\bm{h}_{t}^{2}}=\overleftarrow{\rm{LSTM}}\left(\overleftarrow{\bm{h}_{t+1}^{2}}, \ \bm{h}_{t}^{1}\right),\\
&\bm{h}_{t}^{2}=\left[ \overrightarrow{\bm{h}_{t}^{2}}, \ \overleftarrow{\bm{h}_{t}^{2}} \right].
\end{split}
\end{equation}

Both directions of hidden states at each time step in the first layer are concatenated into $\bm{h}_{t}^{1}$ and fed into the second layer. Two directional hidden states of the second layer are concatenated into $\bm{h}_{t}^{2}$. Then the \emph{t-th} word representation is the concatenation of $\bm{h}_{t}^{1}$ and $\bm{h}_{t}^{2}$, \emph{i.e.}, $\bm{h}_{t}=\left[\bm{h}_{t}^{1}, \bm{h}_{t}^{2}\right]$ $\in{\mathbb{R}^{C_{h}}}$ $\left(t\in{[1,T]}\right)$. The encoded word representation $\bm{h}_{t}$ contains the information not only from the \emph{t-th} word but also from the contextual words before and after the \emph{t-th} word. Afterward, two separate FC layers followed by softmax layers are used to compute the attention weight $\gamma{_{t}^{m}}$ ($m\in{\{actor, action\}}$) on each word,
\begin{equation}
\gamma{_{t}^{m}}=\frac{exp\left(W_{m}\bm{h}_{t}\right)}{\sum_{i=1}^{T}exp\left(W_{m}\bm{h}_{i}\right)},
\end{equation}
where the attention weight $\gamma{_{t}^{m}}$ denotes the probability of the \emph{t-th} word belong to the actor or action. Finally, the $\gamma{_{t}^{m}}$ is applied to the embedding vector sequence $\{\bm{e}_{t}\}_{t=1}^T$ to derive the actor-related and action-related query representations,
\begin{equation}
\bm{q}^{m}=\sum\nolimits_{t=1}^{T}\gamma{_{t}^{m}}\cdot{\bm{e}_{t}}.
\end{equation}

According to the above, we adopt the attention mechanism to adaptively learn the related information about the actor and action. The learning of attention weights is weakly supervised by collaborating with the actor and action modules.

\subsection{Actor and Action Matching in the Modular Network}
\label{section: modular_network}

To segment the actor according to descriptions about actor and its action, we first localize the tube involving the actor and its action in the video clip. We formulate this problem as a matching process in two modules, namely, the actor module and the action module. The former module is applied to localize the tube which involves the actor referred by the given query. It will compute a set of matching scores $s(t_{k} | \bm{q}^{actor})$ ($k\in{[1,K]}$) to measure the similarity between the actor-related query representation $\bm{q}^{actor}$ and the \emph{k}-th tube $t_{k}$. The later module is used to localize the tube which involves the action described by the given query. This module computes a set of matching scores $s(t_{k} | \bm{q}^{action})$ ($k\in{[1,K]}$) to measure the similarity between the action-related query representation $\bm{q}^{action}$ and the \emph{k}-th tube $t_{k}$. The combined matching score $s_{k}=s(t_{k} | \bm{q}^{actor})$ $ + \ s(t_{k} | \bm{q}^{action})$ determines the similarity between the textual query $q$ and the \emph{k}-th tube $t_{k}$. Finally, the tube with the highest score is selected as the target tube which contains the actor and action referred by the textual query. The detailed design of two modules will be discussed in the following parts.

\subsubsection{The Actor Module}
\label{section: actor_module}

\begin{figure}
\centering
\includegraphics[width=\linewidth]{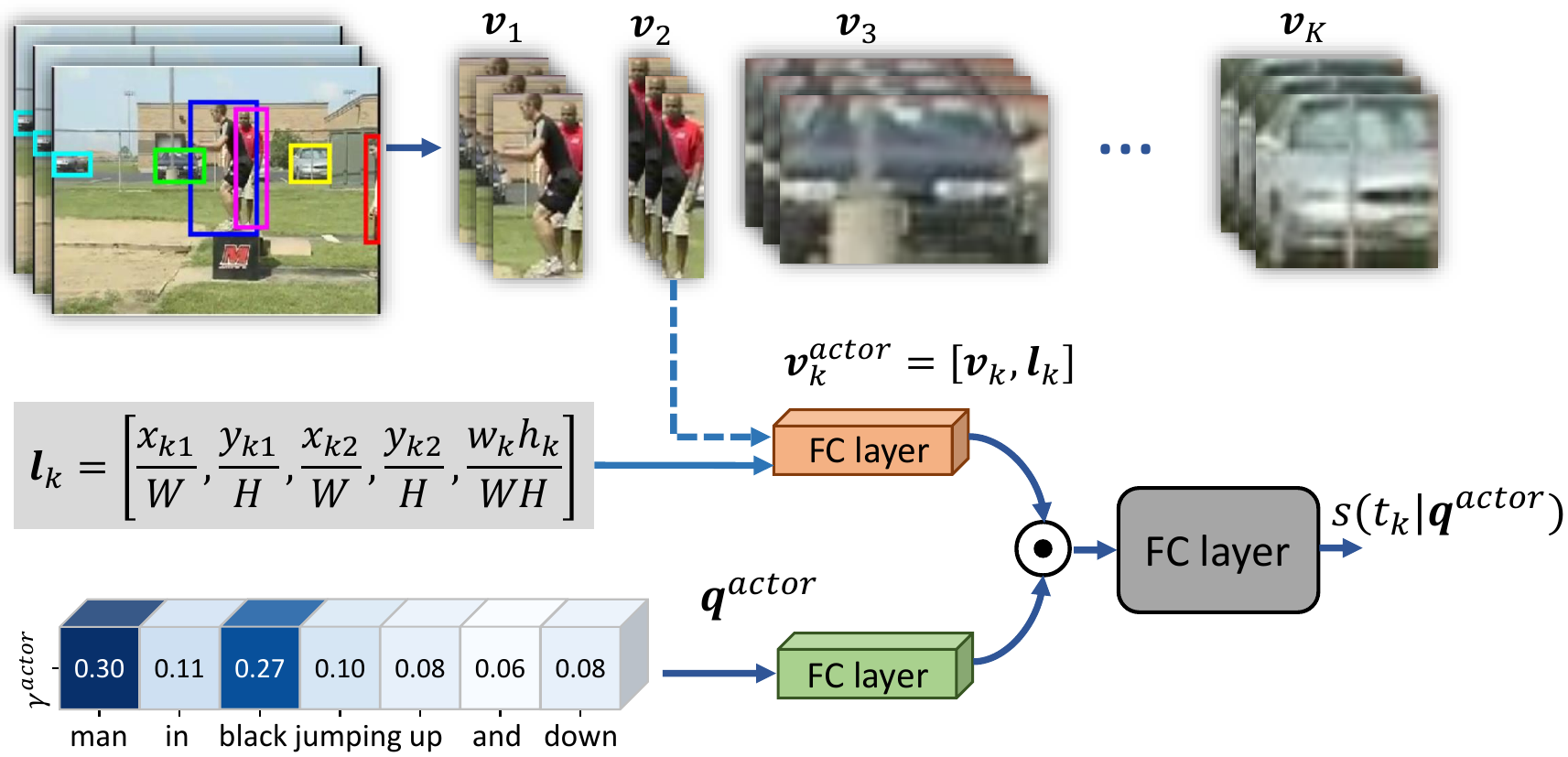}
%\vspace{-0.2cm}
\caption{Illustration of the actor module in AAMN.}
% This module integrates \emph{k}-th appearance representation with actor-related query representation for the actor matching.
\label{fig:actor module}
\vspace{-0.4cm}
\end{figure}

As human usually tends to describe a specific actor according to appearance and location information in videos, we design the actor module to identify the query-specified actor through appearance features from  RGB stream in AAMN. The actor module is shown in Fig. \ref{fig:actor module}. We measure the similarity score between the actor-related query representation $\bm{q}^{actor}$ and feature vector $\bm{v}_{k}^{actor}$ of each tube. The tube representation captures appearance information but lacks necessary location information to distinguish those actors who have similar appearance information. Therefore, following the previous work \cite{yu2016modeling}, we introduce a 5-dimensional location feature $\bm{l}_{k}=\left[ \frac{x_{tl}}{W},\frac{y_{tl}}{H},\frac{x_{br}}{W},\frac{y_{br}}{H},\frac{w\cdot{h}}{W\cdot{H}}\right]$ in the middle frame of the clip, where $(x_{tl},y_{tl},x_{br},y_{br})$ denotes the top-left and bottom-right coordinates, $(w,h)$ and $(W,H)$ are the width and height of the middle proposal in the \emph{k}-th tube and the video frame, respectively. We concatenate the location feature $\bm{l}_{k}$ with the tube representation $\bm{v}_{k}$ as the visual representation of the actor, \emph{i.e.}, $\bm{v}_{k}^{actor}=[\bm{v}_{k}, \bm{l}_{k}] \in{\mathbb{R}^{C_{v}^{'}+5}}$. The visual representation $\bm{v}_{k}^{actor}$ and linguistic representation $\bm{q}^{actor}$ are linearly transformed with FC layers and $l_{2}$-normalized, and then perform element-wise multiplication to integrate two representations. Finally, the integrated representation is fed into one FC layer to predict an unary matching score for the actor as follows:
\begin{equation}
\begin{split}
& \widetilde{\bm{v}}_{k}^{actor}=W_{1}\bm{v}_{k}^{actor}+\bm{b}_{1}, \\
& \widetilde{\bm{q}}^{actor}=W_{2}\bm{q}^{actor}+\bm{b}_{2}, \\
& \bm{x}_{k}^{actor}=\|{\widetilde{\bm{v}}_{k}^{actor}}\|_{2} \odot \|{\widetilde{\bm{q}}^{actor}} \|_{2}, \\
& s_{k}^{actor}=W_{3}\bm{x}_{k}^{actor}+\bm{b}_{3},
\end{split}
\end{equation}
where $\odot$ denotes element-wise multiplication, $\| \cdot \|_2$ means $l_{2}$-normalization, $\left\{ W_{1}, \bm{b}_{1}, W_{2}, \bm{b}_{2}, W_{3}, \bm{b}_{3} \right\}$ are learnable parameters in three FC layers.

%where the $\odot$ means element-wise multiplication, the $W_{1} \in{\mathbb{R}^{d_{0}\times{(d_{2}+5)}}}$, $b_{1} \in{\mathbb{R}^{d_{0}}}$, $W_{2} \in{\mathbb{R}^{1\times{d_{0}}}}$ and $b_{2} \in{\mathbb{R}^{1}}$ are the parameters of two FC layers.

\subsubsection{The Action Module}
\label{section: action_module}

%different from previous works where context information is simply integrated by feature concatenation, we explicitly measure the different "contribution" by leveraging the self-attention technique.

The action matching needs to introduce dynamic motion information as the visual cues for action perception. It is rarely investigated in previous works. Motivated by the previous approaches in \cite{kalogeiton2017joint,ji2018end} where they introduce optical flow as dynamic motion information for action recognition, we introduce another CNN architecture, \emph{i.e.}, Flow stream, which takes a Flow clip as input, to learn motion information for each proposal. We obtain the holistic representation of motion information $\bm{f}_{k}$ ($k\in{\left[1,K\right]}$) for the \emph{k}-th tube $t_{k}$ with the same way as the appearance representation $\bm{v}_{k}$. However, the dynamic motion representation of each tube sometimes is not enough for action inferring. For instance, given a video with the query of \emph{``a man is standing and watching on the behind"}, the actor of \emph{``man"} performs actions of \emph{``standing"} and \emph{``watching"} without obviously spatial displacements. Thus the action cannot be shown from the dynamic motion information. Besides, the tubes with similar motion patterns also affect the result of action matching. To mitigate these problems, we devise a contextual LSTM to model the relationship between the target tube and others, then utilize the information from the contextual LSTM as auxiliary information for the action matching.

\begin{figure}[!t]
\centering
\includegraphics[width=\linewidth]{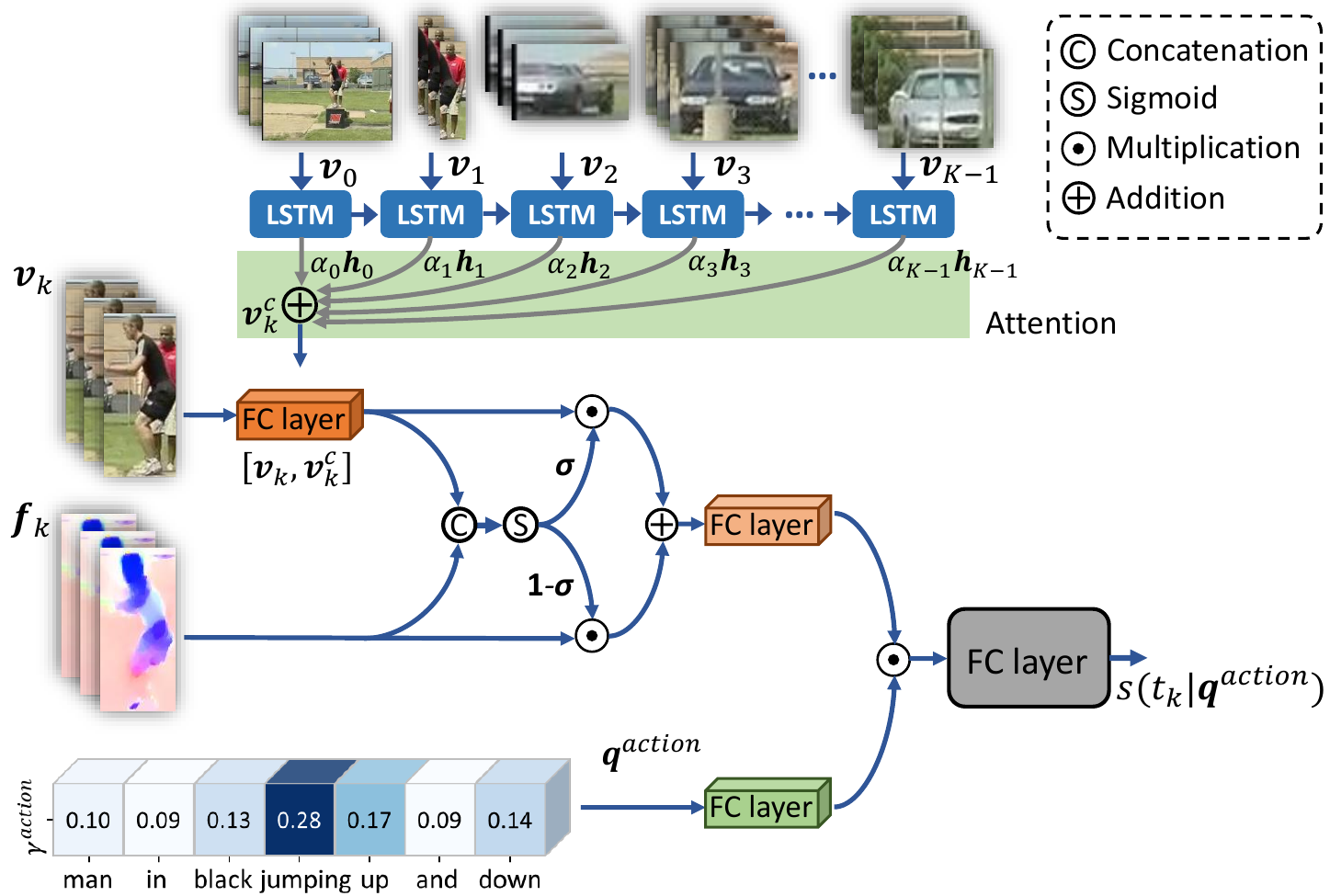}
%\vspace{-0.2cm}
\caption{Illustration of the action module in AAMN.}
% The module integrates visual contextual information, dynamic motion information, and the action-related query representation for action matching.
\label{fig:action module}
\vspace{-0.2cm}
\end{figure}

The proposed action module is shown in Fig. \ref{fig:action module}. This module contains a contextual LSTM network which assists to match the action under the ambiguous motion cues, and a gated fusion network used to weight the importance of motion information and appearance information for action matching. Concretely, the contextual LSTM takes the sequential appearance feature $\bm{v}_{j}\in{\mathbb{R}^{C_{v}^{'}}}$ ($j\in{[1,K]} \ \cap \ j\neq{k}$) of tubes as well as the whole clip feature $\bm{v}_{0} \in{\mathbb{R}^{C_{v}^{'}}}$ as inputs. To obtain $\bm{v}_{0}$, we take the video clip as a large tube. The feature maps of this tube from the last convolutional layer is resized into fixed sized (\emph{e.g.}, $7\times{7}$) via RoIAlign and then fed into two fully connected layers. We aggregate these features using the same way in Equation \ref{equ:avg_pool}. In our model, $\bm{v}_{0}$ represents global contextual information and is arranged in the first place of input sequence, \emph{i.e.}, $\bm{v}_{k}^{seq}=\{\bm{v}_{0},\bm{v}_{1},...,\bm{v}_{k-1},\bm{v}_{k+1},...,\bm{v}_{K-1}\}$ ($k\in{[1,K-1]}$). The tube features are ordered from top to down and left to right according to the proposal locations of the (\emph{L}+1)-th frame in the video clip. Then the sequential features $\bm{v}_{k}^{seq}$ are consecutively fed into LSTM to encode each tube and obtain hidden representations at each time step, \emph{i.e.}, $\bm{h}_{t}^{v} = \rm{LSTM}$$\left(\bm{h}_{t-1}^{v}, \bm{v}_{t} \right) \in{\mathbb{R}^{C_{c}}}$ $\left(t\in{\left[0, K\right]}\right)$. Thereafter, we concatenate all hidden states from each time step $\bm{h}^{v}=\left[ \bm{h}_{0}^{v}, \bm{h}_{1}^{v},...,\bm{h}_{K-1}^{v} \right] \in{\mathbb{R}^{K\times{C_{c}}}}$ and use an attention mechanism to adaptively associate the tubes together to obtain a contextual representation $\bm{v}_{k}^{c} \in{\mathbb{R}^{C_{c}}}$ for the \emph{k}-th tube. The attention is calculated by:
\begin{equation}
\begin{split}
& \bm{y}_{t} = \delta{\left( W_{4}\bm{h}_{t}^{v} + \bm{b}_{4} \right)}, \\
& \alpha{_{t}} = \frac{exp\left(\left(W_{5}\bm{y}_{t} + \bm{b}_{5}\right)^{T}\right)}{\sum_{j=1}^{K}exp\left(\left(W_{5}\bm{y}_{j} + \bm{b}_{5}\right)^{T}\right)}, \\
& \bm{v}_{k}^{c}=\sum\nolimits_{t=1}^{K}\alpha{_{t}}\cdot{\bm{y}_{t}},
\end{split}
\end{equation}
%where $\delta$ is the ReLU function, the $W_{3} \in{\mathbb{R}^{(d_{3}/2) \times{d_{3}}}}$, $b_{3} \in{\mathbb{R}^{d_{3}}}$, $W_{4} \in{\mathbb{R}^{1\times{(d_{3}/2)}}}$ and $b_{4} \in{\mathbb{R}^{1}}$ are the parameters of two FC layers.
where $\delta \left( \cdot \right)$ is ReLU function, $\left\{W_{4}, \bm{b}_{4}, W_{5}, \bm{b}_{5}\right\}$ are learnable parameters in FC layers.

The contextual representation $\bm{v}_{k}^{c}$ of the \emph{k}-th tube is concatenated with appearance representation $\bm{v}_{k}$, \emph{i.e.}, $\widetilde{\bm{v}}_{k}=\left[\bm{v}_{k},\bm{v}_{k}^{c}\right]$ $\in{\mathbb{R}^{C_{v}^{'}+C_{c}}}$. Then $\widetilde{\bm{v}}_{k}$ and the motion representation $\bm{f}_{k} \in{\mathbb{R}^{C_{v}}}$ are fused with a gated way:
\begin{equation}
\begin{split}
& \bm{g}_{k}^{a} = W_{6} \widetilde{\bm{v}}_{k} + \bm{b}_{6}; \ \ \bm{g}_{k}^{b} = W_{7}{\bm{f}}_{k} + \bm{b}_{7},   \\
& \bm{\sigma} \ = \theta \left( W_{8} \left[\bm{g}_{k}^{a},\bm{g}_{k}^{b}\right] + \bm{b}_{8} \right), \\
& \bm{v}_{k}^{action} = \bm{\sigma} \odot \bm{g}_{k}^{a} + \left( \bm{1} - \bm{\sigma} \right) \odot \bm{g}_{k}^{b},
\end{split}
\end{equation}
where $\theta \left( \cdot \right)$ is the sigmoid function which rescales the gate vector $\bm{\sigma} \in{\mathbb{R}^{C_{e}}}$ to $\left[ 0, 1 \right]$. Then the $\bm{\sigma}$ performs element-wise multiplication with sum of $\bm{g}_{i}^{a} \in{\mathbb{R}^{C_e}}$ and $\bm{g}_{i}^{b} \in{\mathbb{R}^{C_e}}$ for weighted fusion them. $\left\{W_{6}, \bm{b}_{6}, W_{7}, \bm{b}_{7}, W_{8}, \bm{b}_{8}\right\}$ are learnable parameters in FC layers.

After obtaining the visual representation of action $\bm{v}_{k}^{action}$ $\in{\mathbb{R}^{C_{e}}}$, we compute the action matching score with the same way as the actor matching score:
\begin{equation}
\begin{split}
& \widetilde{\bm{v}}_{k}^{action}= W_{9}\bm{v}_{i}^{action}+ \bm{b}_{9}, \\
& \widetilde{\bm{q}}^{action} = W_{10}\bm{q}^{action}+\bm{b}_{10}, \\
& \bm{x}_{k}^{action}=\|{\widetilde{\bm{v}}_{k}^{action}}\|_{2} \odot \|{\widetilde{\bm{q}}^{action}} \|_{2}, \\
& s_{k}^{action}=W_{11}\bm{x}_{k}^{action}+\bm{b}_{11},
\end{split}
\end{equation}
where $\left\{W_{9}, \bm{b}_{9}, W_{10}, \bm{b}_{10}, W_{11}, \bm{b}_{11}\right\}$ are corresponding parameters in FC layers.

\subsection{The Actor Segmentation Network}
\label{section: segmentation_network}

Based on the computed matching score $s_{k}$ $\left( i \in{\left[ 1, K\right]} \right)$ above, we can select the target tube with the highest score, which involves the actor and action referred to by the given query. Then we take $2L+1$ feature maps of the target tube into actor segmentation network to predict masks of the actor. The network is a small FCN which takes the fixed size, \emph{e.g.}, $14 \times{14}$, feature maps from RoIAlign layer as input. Similar to Mask R-CNN \cite{he2017mask}, the FCN is comprised of 4 consecutive convolution layers and 1 deconvolutional layer. The first layer consists of 256 $1\times1$ filters to fuse the appearance feature of the tube in RGB stream, and motion feature of the tube in Flow stream. The other three layers are consists of 256 $3\times3$ filters. The final deconvolutional layer up-samples features with factor 2. Since only the (\emph{L}+1)-th frame has mask annotation for input video clip, we choose the (\emph{L}+1)-th feature map of the target tube to train the actor segmentation network with input batch size as 1. During testing, we set input batch as 2\emph{L}+1 and take total feature maps of the target tube to predict the actor masks in the actor segmentation network.

\subsection{Model Learning with Joint Actor-Action Matching and Segmentation}
\label{section: training_model}

During training, AAMN enables joint learning for the actor-action matching and segmentation by minimizing the following combined loss function:
\begin{equation}
\label{equ:total_loss}
L=L_{mat}+\lambda{L_{seg}},
\end{equation}
where $\lambda$ is a hyperparameter to balance the matching loss $L_{mat}$ and the segmentation loss $L_{seg}$. Assuming the ground truth of the actor is contained in the \emph{k-th} region proposal of the (\emph{L+1})-th frame, then $L_{seg}$ will only be computed on the \emph{k}-th mask. $L_{seg}$ is the average binary cross-entropy loss which is defined as:
\begin{equation}
%  \begin{split}
  L_{seg} =  \\
        -\dfrac{1}{h'w'}\sum_{i=1}^{w'}\sum_{j=1}^{h'}(\eta{_{ij}} log(\kappa{_{ij}})+(1-\eta{_{ij}})log(1-\kappa{_{ij}})),
%  \end{split}
\end{equation}
where $h'$ and $w'$ are height and width of the feature map, respectively. $\eta$ is the binary ground-truth label, $\kappa$ is the predicted value of the segmentation mask.

Since the triplet hinge loss has shown its strength in many cross-modal matching problem \cite{faghri2018vse++,niu2019improving}, here we adopt it as the matching loss $L_{mat}$ for the actor and action matching in AAMN. $L_{mat}$ encourages the scores of matched proposals and textual queries to be larger than those of mismatched ones:
\begin{align}
\begin{split}
\label{equ:matching_loss}
L_{mat} = & \sum\nolimits_{q'}{max\left[0, \ \epsilon - s\left(p, \ q\right) + s\left(p, \ \widehat{q}\right)\right]} \\
         +& \sum\nolimits_{p'}{max\left[0, \ \epsilon - s\left(q, \ p\right) + s\left(q, \ \widehat{p}\right)\right]},
\end{split}
\end{align}
where \emph{p} and \emph{q} indicate region proposal and textual query, respectively. $\left(p, q\right)$ and $\left(q, p\right)$ mean the matched proposal and query pairs, and $\left(p, \widehat{q} \right)$ and $\left(q, \widehat{p}\right)$ are the mismatched pairs. $s\left(p, q\right)$ is the computed similarity between \emph{p} and \emph{q}. There are $K-1$ mismatched pairs for each matched pair in our experiments. $\epsilon$ is a margin parameter between matched and mismatched pairs.

During testing, we follow previous works \cite{gavrilyuk2018actor,wang2019asymmetric,mcintosh2018multi} to sample video clips round the annotated frames as inputs. We predict segmentation masks for the selected tube and recover predicted masks into the same resolution as the input clip. Finally, we select the mask in the middle frame (\emph{i.e.}, annotated frame) to evaluate our model.

\section{Experimental Results and Discussions}
\subsection{Datasets and Evaluation Metrics}
%\subsubsection{A2D Sentences}
\textbf{A2D Sentences}. Gavrilyuk \emph{et al.} \cite{gavrilyuk2018actor} augmented Actor-Action Dataset (A2D) with textual queries that describe actors and actions of interest in videos. The original A2D \cite{xu2015can} consists of 7 actor classes which perform one of the 9 action classes. Each video provides around 3 to 5 frames sparse annotations with joint pixel-wise semantic mask and bounding box per actor-action pair. A2D sentences manually annotated 6,655 textual queries which describe actors and actions presented in videos. It is divided into 3,017 and 737 videos for training and testing. Furthermore, McIntosh \emph{et al.} \cite{mcintosh2018multi} annotated actors with bounding boxes in all frames of videos. The extended A2D Sentences makes it available to evaluate the full video segmentation from a textual query.

%\subsubsection{J-HMDB Sentences}
\textbf{J-HMDB Sentences}. J-HMDB Sentences is extended from J-HMDB \cite{jhuang2013towards}, which is a benchmark dataset for action recognition. It comprises of 928 video clips for 21 action classes and provides pixel-wise segmentation mask for the human in action for each frame. There are 928 textual queries are annotated in J-HMDB dataset to describe what action the humans is performing in the video.

%\subsection{Evaluation Metrics}

\textbf{Evaluation Metrics}. We adopt four types of metrics to evaluate our method as prior works \cite{gavrilyuk2018actor,mcintosh2018multi}. For single-frame video segmentation \cite{gavrilyuk2018actor}, each key-frame with a query is regarded as a single sample. While for the full video segmentation \cite{mcintosh2018multi}, each video with a query is considered as a single sample. The Overall Intersection-over-Union (\emph{Overall IoU}) calculates the total intersection area between predictions and ground truth masks divided by total union area accumulated over all test samples, which tends to favor large actors and objects. The \emph{Mean IoU}, which treats large and small actors or objects equally, is calculated as the average over the \emph{IoU} of each testing samples. The precision at different threshold \emph{P@X} ($X \in{\{0.5, 0.6, 0.7, 0.8, 0.9\}}$), which is the percentage of testing samples whose \emph{IoU} higher than specific threshold. The mean average precision (\emph{mAP}) over $\left[0.50:0.05:0.95\right]$ is also computed to measure the segmentation performance.

\subsection{Implementation Details}

% frame size 512x512; frame_num is batch size

In AAMN implementation, we choose Faster R-CNN \cite{ren2015faster} with VGG-16 network as two CNN architectures to extract fixed-size proposal features with RoIAlign layer \cite{he2017mask}. AAMN takes a textual query, a RGB clip and a Flow clip as well as pre-generated proposals as inputs. The frames of the video clip are resized and padded to $512\times{512}$ as previous works \cite{gavrilyuk2018actor,wang2019asymmetric}. The optical flow is a tensor of three channels with x and y coordinates of the flow as well as the flow magnitude \cite{kalogeiton2017joint,ji2018end}. Since the generation of high-quality proposals is important for actor and action localization, we choose Faster R-CNN \cite{ren2015faster} with a backbone of ResNet-101 as an external object detector. This model is trained on MSCOCO dataset \cite{lin2014microsoft} and further finetuned on A2D \cite{gavrilyuk2018actor}. In our experiments, the dimension of $C_{v}$ is 4096, the dimension of $C_{v}^{'}$ is set as 1024. For the textual query encoding, we adopt two-layer Bi-directional LSTM (Bi-LSTM). The dimension of the word embeddings that are input into Bi-LSTM is $C_{e}=300$, and the forward and backward hidden states from Bi-LSTM are $C_{h}=512$. In order to obtain fixed dimensional query representation, long queries are truncated and short queries are padded with zeros. The fixed length of the query is set to 20 in our experiments. Besides, the dimension of hidden states in the contexual LSTM is $C_{c}=512$.

We implement our proposed model with Tensorflow. Two CNN architectures are initialized with pre-trained weights on MSCOCO dataset \cite{lin2014microsoft}. During the model training, we use stochastic gradient descent (SGD) with initial learning rate of 0.0001, momentum of 0.95, weight decay of 0.0005. The learning rate decreases by 10 times at $1.0\times{10^5}$ iterations and the model is trained for $1.5\times{10^5}$ iterations. The balanced parameter $\rho$ in Equation \ref{equ:linking_score} is selected as 1.0. The margin $\epsilon$ is set to 0.1 in the matching loss $L_{mat}$ in Equation \ref{equ:matching_loss}.

\subsection{Ablation Studies}

\subsubsection{Selection of Proposal Number}

The selection of proposal number is critical to model visual region context in the contextual LSTM of the action module. Thus we conduct experiments with various numbers of proposals to analyze the impact of proposals number in our model. We first extract a set of proposals per frame with high confidence utilizing an object detector. Then we select a specific number of proposals and feed them into our model. In our experiments, the number is set to increase from 2 to 12 with an interval of 2. Before evaluating the segmentation results with different proposals number, we first analyze the actor and action localization results with bounding boxes. Here, the precision of localization \emph{P@X} $\left(X \in{\{0.5,0.6,0.7,0.8,0.9\}}\right)$ is evaluated by calculating the IoU ratio between the true bounding box and the top predicted box in annotated frames. If the IoU is larger than the threshold, the prediction is considered as a true positive, otherwise it is counted as a false positive. The total true positive counts are then averaged over all testing samples to obtain \emph{P@X}. Table \ref{table:precision_localization} shows the precision of localization at versus thresholds with various numbers of proposals. It can be seen that the precision increases with increment of proposals number. When the number reaches a certain value, \emph{i.e.}, $K=6$, the precision tends to decline in most metrics. This reveals that the contextual LSTM progressively encodes local features into a discriminate vector representation which contains more details of video content, but too more proposals may bring noises. The segmentation results with different number of proposals are shown in Table \ref{table:precision_segmentation}. Since our model is a two-stage method, the segmentation performance is closely relevant to the localization performance, the best segmentation performance is achieved when the number of proposals equals 6. Thus we set the number of proposals as 6 in our model.

% table 1: number of object proposals with boxes precision
\begin{table}[!tb]
%\small
\begin{threeparttable}
\caption{Analysis of localization precision at different threshold with different proposals number in each video frame}
\centering
\label{table:precision_localization}
\setlength{\tabcolsep}{3.3mm}{
\begin{tabular}{ccccccc}
\toprule
\textbf{Models} & \textbf{P@0.5} & \textbf{P@0.6} & \textbf{P@0.7} & \textbf{P@0.8} & \textbf{P@0.9} \\
\midrule
K =$ \ $ 2  &67.5    &63.8    &58.7   &48.5    &20.1\\
K =$ \ $ 4  &\textbf{69.1}    &65.2   &59.6    &\textbf{49.1}    &20.7 \\
K =$ \ $ 6  &69.0	 &\textbf{66.1}   &\textbf{60.2}	  &48.7	   &\textbf{21.5} \\
K =$ \ $ 8  &68.2    &65.5   &59.7    &48.1    &20.9 \\
K = 10      &67.9    &65.0   &59.4    &47.9    &21.0 \\
K = 12      &67.2	 &63.5	 &57.9	  &46.5	   &19.3 \\

\bottomrule
\end{tabular}}
\end{threeparttable}
\vspace{-0.3cm}
\end{table}

% table 2: number of object proposals
\begin{table}[!tb]
%\small
%\begin{center}
\begin{threeparttable}
%\fontsize{6.5}{8}\selectfont
\caption{The analysis of segmentation results with different proposal number}
\centering
\label{table:precision_segmentation}
\setlength{\tabcolsep}{0.9mm}{
\begin{tabular}{ccccccccc}

\toprule
\multirow{2}{*}{\textbf{Models}}&\multicolumn{5}{c}{\textbf{Overlap}}&\textbf{mAP}&\multicolumn{2}{c}{\textbf{IoU}}\\
\cmidrule(lr){2-6}
\cmidrule(lr){7-7}
\cmidrule(lr){8-9}
& P@0.5 & P@0.6 & P@0.7 & P@0.8 & P@0.9 & 0.5:0.95 & Overall & Mean \\
\midrule
K = \ 2     & 66.9  &60.6   &50.7   &26.5   &1.7    &36.8   &59.9   &53.6 \\
K = \ 4     & 68.7  &\textbf{64.6}  &53.6   &30.4   &3.1    &39.5   &62.9   &\textbf{55.7} \\
K = \ 6     & \textbf{68.9}	&64.2	&\textbf{54.5}	&\textbf{32.4}	&\textbf{3.4}	&\textbf{41.2}	&\textbf{63.4}	&55.6 \\
K = \ 8     & 67.7  &63.2   &53.1   &31.6   &2.9    &40.6   &62.7   &54.8 \\
K = 10      & 67.1  &62.9   &52.6   &29.3   &2.2    &39.9   &60.9   &53.3 \\
K = 12      & 66.6	&61.8	&50.5	&27.8	&2.1	&37.2	&58.8	&52.7 \\

\bottomrule
\end{tabular}}
\end{threeparttable}
%\end{center}
\vspace{-0.3cm}
\end{table}

% table 3: different frames for a clip
\begin{table}[!tb]
%\small
\begin{threeparttable}
%\begin{center}
%\fontsize{6.5}{8}\selectfont
\caption{The impact of input frame number}
\label{table:frame_number}
\centering
\setlength{\tabcolsep}{1.0mm}{
\begin{tabular}{lcccccccc}

\toprule
\multirow{2}{*}{\textbf{Models}}&\multicolumn{5}{c}{\textbf{Overlap}}&\textbf{mAP}&\multicolumn{2}{c}{\textbf{IoU}}\\
\cmidrule(lr){2-6}
\cmidrule(lr){7-7}
\cmidrule(lr){8-9}
& P@0.5 & P@0.6 & P@0.7 & P@0.8 & P@0.9 & 0.5:0.95 & Overall & Mean \\
\midrule

L = 0      & 68.1	& 62.9	  & 52.3	& 29.4	 & 2.8	 & 39.6	   & 61.7	 & 55.2 \\
L = 2      & 68.4	& 64.1	  & 53.2	& 30.1	 & 2.9	 & 39.9	   & 62.7	 & 55.3 \\
L = 4      & \textbf{68.9}	& 64.2	   & \textbf{54.5}	& \textbf{32.4}	 & \textbf{3.4}	 & \textbf{41.2}	   & \textbf{63.4}	 & \textbf{55.6} \\
L = 8      & \textbf{68.9}	& \textbf{64.4}	  & 54.2	& 31.9	 & 3.2	 & 40.5	   & 62.8	 & 55.5 \\

\bottomrule
\end{tabular}}
%\end{center}
\end{threeparttable}
\vspace{-0.1cm}
\end{table}

% table 6: effect of different components
\begin{table*}[!tb]
%\begin{center}
\begin{threeparttable}
\caption{Ablation study on A2D Sentences}
\label{table:Ablation_study}
\centering
\setlength{\tabcolsep}{3.8mm}{
\begin{tabular}{lccccccccc}

\toprule
\multirow{2}{*}{\textbf{Models}}&\multicolumn{5}{c}{\textbf{Overlap}}&\textbf{mAP}&\multicolumn{2}{c}{\textbf{IoU}}\\
\cmidrule(lr){2-6}
\cmidrule(lr){7-7}
\cmidrule(lr){8-9}
                                                & P@0.5      & P@0.6     & P@0.7    & P@0.8    & P@0.9    & 0.5:0.95    & Overall     & Mean \\
\midrule
Baseline Model                                  & 63.3	      &56.7       &39.5      &17.2      &1.1       &31.9          &55.2        &48.5 \\
AAMN \emph{w/o} Actor Module                    & 48.0        &44.3       &36.8      &20.7      &1.9       &27.8          &45.7        &39.7 \\
AAMN \emph{w/o} Action Module                   & 63.7        &58.7       &42.5      &22.3      &1.4       &33.7          &58.9        &51.8 \\
AAMN \emph{w/o} Location Feature                & 68.2	      & 63.7	  &53.4	     & 31.8     & 2.7      & 39.5	      &62.4	       &54.7 \\
AAMN \emph{w/o} Contextual LSTM                 & 67.9	      & 62.5	  &51.8      & 27.8     & 2.0      & 37.8         &60.3        &53.6 \\
AAMN \emph{w/o} Attention                       & 68.3	      & 63.4	  &52.8	     & 31.7     & 2.1      & 39.9	      &61.5	       &54.8 \\
AAMN \emph{w/o} Gated Fusion                    & 68.1        & 62.7	  &52.6	     & 29.7     & 2.2      & 38.9	      &61.7	       &53.8 \\
\cmidrule(lr){1-9}
AAMN \emph{w/o} Flow Stream                     & 65.2	      & 58.0	  & 44.5	 & 20.8	    & 1.6	   &34.8	      &58.8	       &51.3 \\
AAMN \emph{w/o} RGB Stream                      & 55.1	      & 43.3	  & 22.7	 & 4.5	    & 0.1	   &22.2	      &46.5	       &42.0 \\
AAMN (FCN) \emph{w/o} Optical Flow              & 67.5        & 61.9      & 49.1     & 26.4     & 2.0      &38.0          &61.8        &53.9 \\
AAMN (Action Module) \emph{w/o} Optical Flow    & 67.1        & 62.2	  &51.6	     & 29.3     & 1.6      & 38.0	      &60.9        &53.3 \\
AAMN (Action Module) \emph{w/o} RGB             & 65.9	      & 61.7	  &52.2	     & 31.4     & 1.4      & 37.8	      &59.8	       &52.5 \\
AAMN (Actor Module) \emph{w/} Optical Flow      & 68.0	      & 63.1	  &53.0	     & 30.0     & 2.4      & 39.0	      &61.7	       &54.3 \\
%\cmidrule(lr){1-9}
%AAMN w/o Detector Finetuning                    & 64.2        & 59.4      & 49.1     & 27.4     & 2.3      &37.4          &57.4        &51.7 \\
%\cmidrule(lr){1-9}
AAMN (Full Model)                    & \textbf{68.9} &\textbf{64.2} &\textbf{54.5} &\textbf{32.4} &\textbf{3.4} &\textbf{41.2} &\textbf{63.4} &\textbf{55.6} \\
\bottomrule
\end{tabular}}
\end{threeparttable}
%\end{center}
%\vspace{-0.2cm}
\end{table*}

% table 4: different approaches for linking scores computation
\begin{table}[!tb]
%\small
\begin{threeparttable}
%\begin{center}
%\fontsize{6.5}{8}\selectfont
\caption{The Impact of Optical Flow on the Semantic Similarity}
\label{table:linking_scores}
\centering
\setlength{\tabcolsep}{0.7mm}{
\begin{tabular}{lcccccccc}

\toprule
\multirow{2}{*}{\textbf{Features}}&\multicolumn{5}{c}{\textbf{Overlap}}&\textbf{mAP}&\multicolumn{2}{c}{\textbf{IoU}}\\
\cmidrule(lr){2-6}
\cmidrule(lr){7-7}
\cmidrule(lr){8-9}
& P@0.5 & P@0.6 & P@0.7 & P@0.8 & P@0.9 & 0.5:0.95 & Overall & Mean \\
\midrule

Flow      & 67.7 	& 62.0	  & 49.4	&  23.7 	 & 0.8	   & 37.4      & 61.1	   & 53.5    \\
RGB       & \textbf{68.9}	& \textbf{64.2}	   & \textbf{54.5}	& \textbf{32.4}	 & \textbf{3.4}	 & \textbf{41.2}	   & \textbf{63.4}	 & \textbf{55.6} \\
Flow+RGB  & 68.4  	& 64.0 	  & 54.0	& 32.4    & 2.6     & 40.7	   & 62.0   & 55.2  \\

\bottomrule
\end{tabular}}
%\end{center}
\end{threeparttable}
\vspace{-0.1cm}
\end{table}

% table 5: average aggregation
\begin{table}[!tb]
%\small
\begin{threeparttable}
%\begin{center}
%\fontsize{6.5}{8}\selectfont
\caption{Comparison of two temporal aggregation strategies}
\label{table:temporal_aggregation}
\centering
\setlength{\tabcolsep}{0.9mm}{
\begin{tabular}{lcccccccc}

\toprule
\multirow{2}{*}{\textbf{Models}}&\multicolumn{5}{c}{\textbf{Overlap}}&\textbf{mAP}&\multicolumn{2}{c}{\textbf{IoU}}\\
\cmidrule(lr){2-6}
\cmidrule(lr){7-7}
\cmidrule(lr){8-9}
& P@0.5 & P@0.6 & P@0.7 & P@0.8 & P@0.9 & 0.5:0.95 & Overall & Mean \\
\midrule

Average    & \textbf{69.1}	& 63.4	  & 53.0	& 30.4 	 & 2.8	   & 40.7      & 62.8	   & \textbf{55.7}   \\
Ours       & 68.9	& \textbf{64.2}	   & \textbf{54.5}	& \textbf{32.4}	 & \textbf{3.4}	 & \textbf{41.2}	   & \textbf{63.4}	 & 55.6 \\

\bottomrule
\end{tabular}}
%\end{center}
\end{threeparttable}
\vspace{-0.1cm}
\end{table}

\subsubsection{Impact of Input Frame Number}

The AAMN takes RGB and Flow clips as inputs, where each clip contains $2L+1$ frames. To investigate the impact of frame number in the clip, we perform experiments with different \emph{L}, \emph{i.e.}, \emph{L} = \{0, 2, 4, 8\}, the corresponding frames are \{1, 5, 9, 17\} in a clip. The experimental results are shown in Table \ref{table:frame_number}. We observe that the single frame input, \emph{i.e.}, \emph{L} = 0, has the worst performance in all metrics. By increasing the number of input frames, the performance is increased, especially in \emph{P@0.6} and \emph{P@0.7}. It demonstrates that the aggregation of temporally contextual information is helpful for actor and action matching in two modules, and leads to the improvement of actor segmentation. The best results are achieved under most metrics when the frame number is 9. The performance drops by further increasing the frame number. The possible reason is that proposals of the actor or object cannot link well by computing similarity of spatial locations over a long frames sequence, irrelevantly temporal information is aggregated into the annotated frame. Therefore, we select the frame number as 9, \emph{i.e.}, \emph{L} = 4, in our experiments.

\subsubsection{Study of Ablation Models}
To systematically evaluate the relative contributions of different components in AAMN, we design seven ablation models as shown in Lines 1-7 of Table \ref{table:Ablation_study}. a) ``Baseline Model" utilizes concatenation of the appearance feature $\bm{v}_{i}$ and the motion feature $\bm{f}_{i}$ of the \emph{i}-th tube as the visual representation, the mean pooling of embedded word sequence as a language representation. The matching score is computed using the same way as the actor matching score. b) ``\emph{w/o} Actor Module" means that AAMN only contains the action module. c) ``\emph{w/o} Actor Module" means that AAMN only contains the action module. d) ``\emph{w/o} Location Feature" indicates that the actor module without the location feature $\bm{l}_{i}$. e) ``\emph{w/o} Contextual LSTM" indicates that the action module without the contextual LSTM. f) ``\emph{w/o} Attention" refers to using element-wise mean pooling over the concatenated hidden states $\bm{h}^{v}$ in the contextual LSTM to obtain the contextual representation $\bm{v}_{i}^{c}$. g) ``\emph{w/o} Gated Fusion" refers to using concatenation instead of gated fusion to merge the appearance feature $\widetilde{\bm{v}}_{i}$ and the motion feature $\bm{f}_{i}$ in the action module. As a comparison, the ``Full Model" is shown in the last Line of Table \ref{table:Ablation_study}.

From the experimental results in Lines 1-7 and Line 14 of Table \ref{table:Ablation_study}, we can observe that: i) Benefiting from the modular design, the ``Full Model" outperforms the ``Baseline Model" by a large margin in terms of different metrics. ii) The performance of individual modules is close to or slightly worse than the performance of ``Baseline Model", while the combination of them can significantly improve the performance. This demonstrates that two modules can collaborate and facilitate each other for the actor and action matching. iii) Modeling the contextual information using the contextual LSTM with attention is helpful to promote the action matching and further improve the segmentation performance. iv) Compared with the concatenation way to fuse the appearance feature $\widetilde{\bm{v}}_{i}$ and the motion feature $\bm{f}_{i}$, the gated fusion way can learn to weight the importance of the motion feature $\bm{f}_{i}$ and the appearance feature $ \widetilde{\bm{v}}_{i}$ for accurately matching the described action in a textual query.

\subsubsection{Impact of Optical Flow on Different Components}

The impact of optical flow have been studied when we add optical flow features in different components of AAMN, the corresponding results are shown in Lines 8-13 of Table \ref{table:Ablation_study}. a) ``\emph{w/o} Flow Stream" denotes that AAMN only contains RGB stream and the motion feature $\bm{f}_{i}$ is replaced with the appearance feature $\bm{v}_{i}$ in action module. b) ``\emph{w/o} RGB Stream" means that AAMN only contains flow stream and the appearance feature $\bm{v}_{i}$ is replaced with the motion feature $\bm{f}_{i}$ in two modules. c) ``(FCN) \emph{w/o} Optical Flow" indicates that AAMN predicts the segmentation mask from the appearance features $\bm{v}_{i}$. d) ``(Action Module) \emph{w/o} Optical Flow" denotes that AAMN only uses the appearance feature $\widetilde{\bm{v}}_{i}$ for action matching in the action module. e) ``(Action Module) \emph{w/o} RGB" means that AAMN only utilizes the motion feature $\bm{f}_{i}$ for action matching in the action module. f) ``(Actor Module) \emph{w/} Optical Flow" refers to adding the motion feature $\bm{f}_{i}$ into the actor module.

From the experimental results in Lines 8-14 of Table \ref{table:Ablation_study}, we can conclude that: i) The two-stream framework performs better than single stream framework, because the appearance feature $\bm{v}_{i}$ is essential for actor matching and the motion feature $\bm{f}_{i}$ plays a key role for action matching. ii) The combination of the motion feature $\bm{f}_{i}$ and the appearance feature $\bm{v}_{i}$ can improve the segmentation performance in AAMN. iii) The individual appearance feature $\widetilde{\bm{v}}_{i}$ or motion feature $\bm{f}_{i}$ performs worse than their combination for action matching, since the appearance feature $\widetilde{\bm{v}}_{i}$ encoded region contextual information can distinguish those actions which have ambiguous motion cues. iv) The motion feature $\bm{f}_{i}$ cannot promote the actor matching, this may be because the input of two modules are same, the modular network cannot learn discriminative information for the actor and action matching.

We also investigate the impact of optical flow on the semantic similarity of linking score in Equation \ref{equ:linking_score}. Specifically, we generate tubes by calculating the semantic similarity using visual features from Flow stream, RGB stream, and combination of two streams. Experimental results are summarized in Table \ref{table:linking_scores}. It can be observed that our model achieves the best performance by calculating the semantic similarity only using visual features from RGB stream, while the performance degrades obviously using visual features from Flow stream or combination of two streams. This is reasonable because we expect to link objects in consecutive frames by capturing invariant features about these objects. Specifically, visual features from RGB stream learn appearance information about objects. Such information contains visual cues about categories, colors, and shapes of objects. Thus, appearance information is usually invariant between adjacent frames. However, visual features from Flow stream mainly learn dynamic motion information performed by objects. For each object, the learned information is not stable in single frames for different actions, view angles, movement speeds, and occlusions. Lots of works \cite{kong2022human} usually aggregate multiple frames of visual features from Flow stream for action recognition. Calculating the semantic similarity with such dynamic motion information is not accurate to link objects. Therefore, we only use visual features from RGB stream to calculate the semantic similarity in our work.

\subsubsection{Impact of Temporal Aggregation Strategy}
Due to sparse annotations of A2D dataset, we generate the tube representation by aggregating visual features from adjacent frames into the annotated frame (\emph{i.e.}, the middle frame) to train our model. Except the temporal aggregation strategy in Equation \ref{equ:avg_pool}, we also explored another aggregation strategy by directly averaging visual features over all frames of the clip. Experimental results are summarized in Table \ref{table:temporal_aggregation}. It can be observed that the strategy in Equation \ref{equ:avg_pool} obtains better results than averaging visual features over all frames in most metrics. This is because visual features from the annotated frame are more discriminative for actor and action matching, and assigning larger weights to visual features from annotated frame can improve the localization precision of the actor.

\subsubsection{Analysis of Learning Strategy}
Our AAMN is a two-stage method which first localizes the actor-/action-related tube and then performs segmentation within the localized tube. Thus we can choose two learning strategies for AAMN: jointly learn actor-action matching and segmentation at same time, or separate learn them one-by-one. To compare with our main joint learning strategy, we perform two separate learning experiments, \emph{i.e.}, ``M + S" and ``S + M". The ``M + S" means we first train the matching branch until convergence and then train the FCN branch for the actor segmentation. The ``S + M" denotes we first train the FCN branch until convergence and then train the matching branch. The experimental results with different learning strategies are shown in Table \ref{table:learning_strategy}. Note that the actor-action matching and segmentation branches share the same backbone of feature extractor, therefore when separately trained, the final AAMN may be bias to the actor-action matching or segmentation. By comparing the results in Table \ref{table:learning_strategy}, we can conclude that joint learning is helpful to avoid subtask biases and achieves best performance.

% table: step training and joint training
\begin{table}[!bt]
%\small
\begin{threeparttable}
%\begin{center}
%\fontsize{6.5}{8}\selectfont
\caption{The analysis of different training strategies}
%\vspace{-0.5cm}
\label{table:learning_strategy}
\centering
\setlength{\tabcolsep}{1.0mm}{
\begin{tabular}{lcccccccc}

\toprule
\multirow{2}{*}{\textbf{Models}}&\multicolumn{5}{c}{\textbf{Overlap}}&\textbf{mAP}&\multicolumn{2}{c}{\textbf{IoU}}\\
\cmidrule(lr){2-6}
\cmidrule(lr){7-7}
\cmidrule(lr){8-9}
& P@0.5 & P@0.6 & P@0.7 & P@0.8 & P@0.9 & 0.5:0.95 & Overall & Mean \\
\midrule
M + S           &66.2	&55.2	&34.3	&9.3	&0.4	&29.8	&55.7	&50.0\\
S + M           &65.4	&58.5	&41.9	&15.9	&1.6	&33.1	&57.9	&50.8 \\
Joint         &\textbf{68.9}	&\textbf{64.2}	&\textbf{54.5}	&\textbf{32.4}	&\textbf{3.4}	&\textbf{41.2}	 &\textbf{63.4}	&\textbf{55.6}\\

\bottomrule
\end{tabular}}
%\end{center}
\end{threeparttable}
\vspace{-0.3cm}
\end{table}

% table : hyper-parameters
\begin{table}[!bt]
%\small
\begin{threeparttable}
%\begin{center}
%\fontsize{6.5}{8}\selectfont
\caption{The analysis of hyper-parameter $\lambda$}
%\vspace{-0.5cm}
\label{table:hyper-parameter}
\centering
\setlength{\tabcolsep}{1.2mm}{
\begin{tabular}{lcccccccc}

\toprule
\multirow{2}{*}{\textbf{$\rm{\lambda}$}}&\multicolumn{5}{c}{\textbf{Overlap}}&\textbf{mAP}&\multicolumn{2}{c}{\textbf{IoU}}\\
\cmidrule(lr){2-6}
\cmidrule(lr){7-7}
\cmidrule(lr){8-9}
& P@0.5 & P@0.6 & P@0.7 & P@0.8 & P@0.9 & 0.5:0.95 & Overall & Mean \\
\midrule

0.01        & 59.9	& 44.1	& 20.2	& 2.9	& 0	    & 21.9	& 51.8	& 46.3 \\
0.1         & 64.8	& 54.5	& 35.2	& 10.5	& 0.7	& 29.7	& 56.9	& 49.7 \\
1.0         & 67.0	& 60.7	& 45.0	& 21.8	& 1.2	& 35.7	& 59.5	& 52.9 \\
2.5         & 67.7	& 61.6	& 48.6	& 25.0	& 2.5	& 37.5	& 60.7	& 53.9 \\
5.0         &\textbf{68.9}	&\textbf{64.2}	&\textbf{54.5}	&\textbf{32.4}	&\textbf{3.4}	&\textbf{41.2}	 &\textbf{63.4}	&\textbf{55.6} \\
7.5         & \textbf{68.9}	& 64.1	& 53.8	& 30.6	& 2.3	& 40.5	& 63.0	& \textbf{55.6} \\
10.0        & 68.2	& 63.2	& 53.3	& 30.6	& 2.2	& 40.2	& 62.0	& 55.1 \\
12.5        & 68.3	& 63.5	& 53.1	& 28.9	& 1.6	& 38.8	& 61.3	& 54.2 \\
15.0        & 67.4	& 62.4	& 51.3	& 28.0	& 1.9	& 37.7	& 60.1	& 53.5 \\

\bottomrule
\end{tabular}}
%\end{center}
\end{threeparttable}
\vspace{-0.3cm}
\end{table}

\begin{figure}[!tb]
\centering
\includegraphics[width=\linewidth]{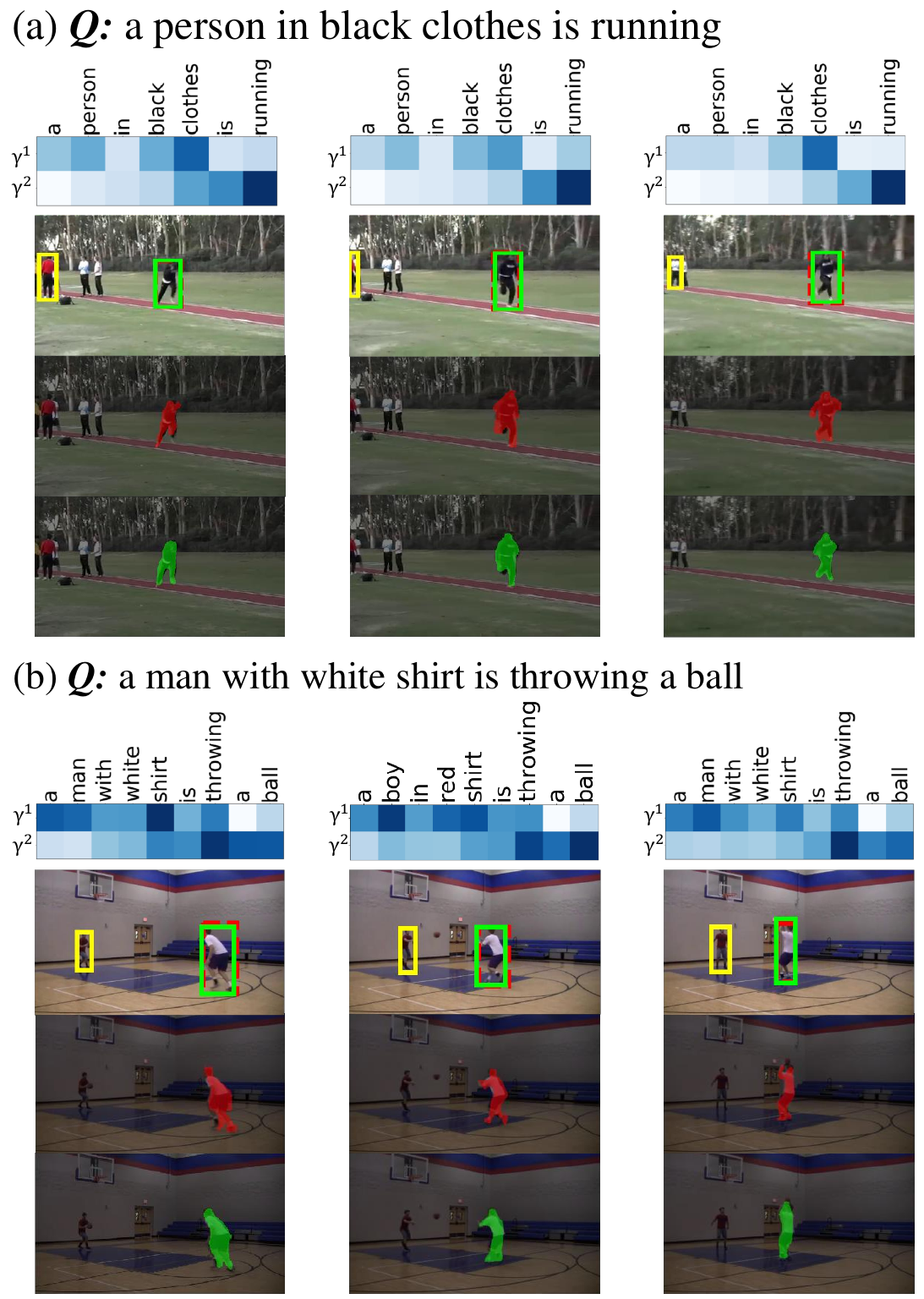}
\caption{Visualization of word attention weights and segmentation results on A2D Sentences. For each example, we sample three annotated frames from a video. The first row shows the learned attention weights for the textual query, the second row shows the localization results in annotated frames, the third row shows the the ground truth masks (red), the last row shows the predicted masks (green) for the referred actor.}
\label{fig:attn_visualization}
\vspace{-0.4cm}
\end{figure}

\subsubsection{Analysis of Hyper-parameter}
In Equation \ref{equ:total_loss}, a hyper-parameter $\lambda$ is introduced to balance the matching loss $L_{mat}$ and the segmentation loss $L_{seg}$ in joint learning process. $\lambda=1.0$ means $L_{mat}$ plays an equally important role as $L_{seg}$. Here we conduct experiments to analyze the impacts of $\lambda$. We vary $\lambda$ from 0.001 to 15.0 in experiments and present corresponding results in Table \ref{table:hyper-parameter}. It can be found that the segmentation performance is improved with increasing of $\lambda$, and reaches the best performance at most evaluation metrics when $\lambda=5.0$. After that, the performance tends to descend when $\lambda$ keeps increasing.

\subsubsection{Analysis of Word Attention Weights}
As the lack of word-level annotations, the language attention learning is weakly supervised by collaborating with the actor and action modules. According to the role of each word plays in two modules, the related information about the actor and action is adaptively learned from the textual query. Since it is hard to quantitatively evaluate the accuracy of learned word attention weights, we visualize word attention weights and corresponding segmentation results on key frames of two video from A2D Sentences in Fig. \ref{fig:attn_visualization}.

In Fig. \ref{fig:attn_visualization}, the first row shows the visualization of learned word attention weights for the textual query, where $\gamma^{1}$ and $\gamma^{2}$ denote word attention weights corresponding to the actor module and action module, respectively. The darker color indicates larger value of weight. We can observe that the language attention model enables to attend to right words that are relevant to the actor and action respectively. Specifically, $\gamma^{1}$ has higher values on those word that are nouns and attribute words referring to the actor. While $\gamma{2}$ obtains higher values on those words that are verbs or verb-related phrases. In the second row, it is obvious that our model can accurately localize the actor with green bounding boxes in different frames. This can be attributed to the modular design in our model. The predicted bounding box from the detector is very close to the ground truth bounding box (red dashed line). The yellow bounding box is the most relevant contextual region which has the highest attention weight in the contextual LSTM of the action module. The third row shows the ground truth segmentation masks of the actor in key frames. The last row shows the predicted segmentation mask of the referred actor with our AAMN. We can observe that our model is able to predict accurate binary masks for the referred actor in key frames.

% table : Ñ¡²»Í¬µÄÖ¡ £¨2L+1£©
\begin{table}[!tb]
%\small
\begin{threeparttable}
%\begin{center}
%\fontsize{6.5}{8}\selectfont
\caption{Comparison with random attention weights.}
\label{table:random_weights}
\centering
\setlength{\tabcolsep}{0.35mm}{
\begin{tabular}{lcccccccc}

\toprule
\multirow{2}{*}{\textbf{Weights}}&\multicolumn{5}{c}{\textbf{Overlap}}&\textbf{mAP}&\multicolumn{2}{c}{\textbf{IoU}}\\
\cmidrule(lr){2-6}
\cmidrule(lr){7-7}
\cmidrule(lr){8-9}
& P@0.5 & P@0.6 & P@0.7 & P@0.8 & P@0.9 & 0.5:0.95 & Overall & Mean \\
\midrule

Random             &65.2 	&60.0	&49.2	&26.8 	 &1.3    &36.5	   &58.3	 &51.8  \\
Prediction (Ours)  & \textbf{68.9}	& \textbf{64.2}	   & \textbf{54.5}	& \textbf{32.4}	 & \textbf{3.4}	 & \textbf{41.2}	   & \textbf{63.4}	 & \textbf{55.6} \\

\bottomrule
\end{tabular}}
%\end{center}
\end{threeparttable}
\vspace{-0.1cm}
\end{table}

\begin{figure}[!tb]
\centering
\includegraphics[width=\linewidth]{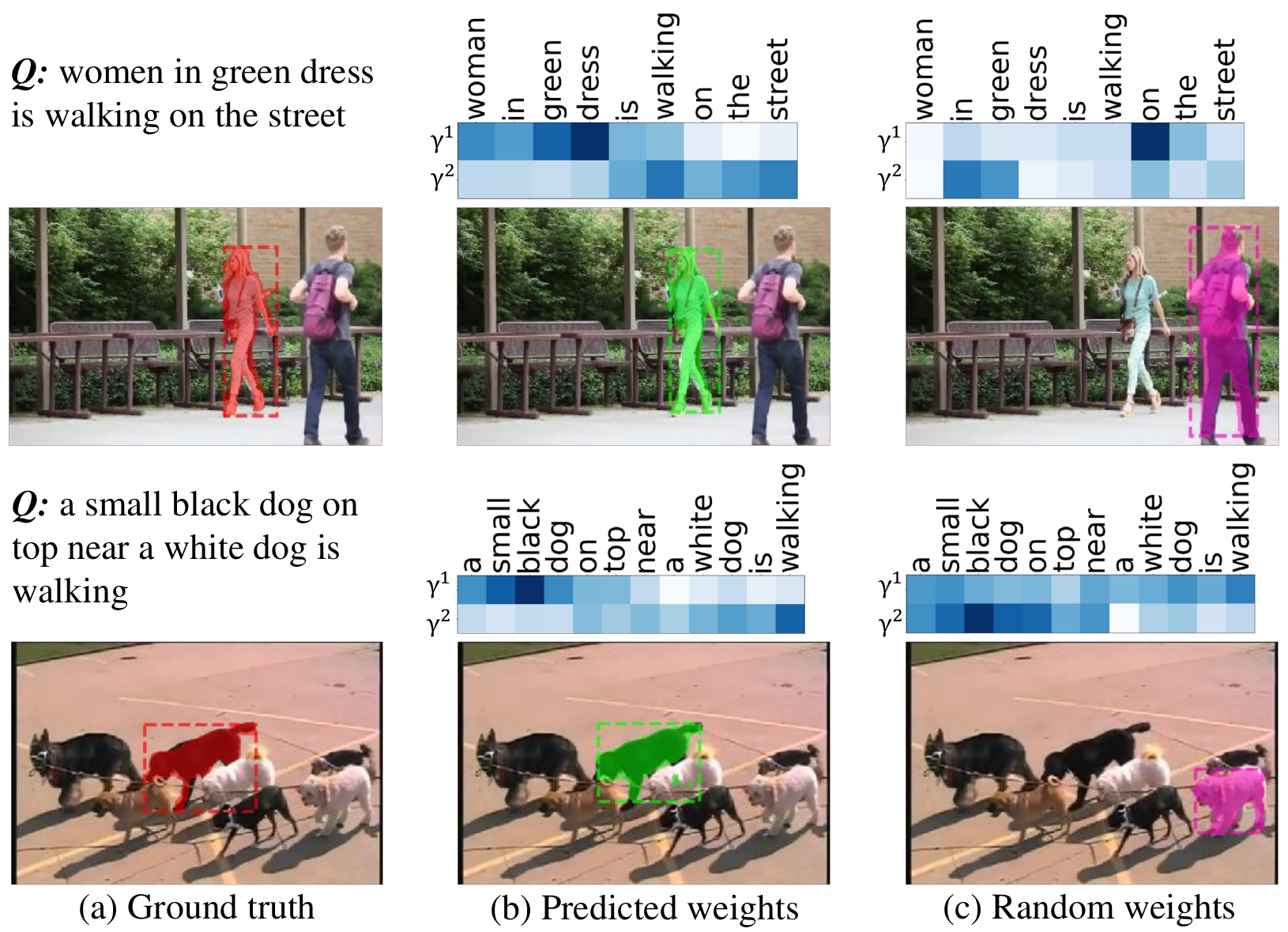}
\caption{Visualization of experimental results with predicted and random attention weights. (a) The ground truth. (b) Segmentation results with predicted word attention weights. (c) Segmentation results with random word attention weights.}
\label{fig:random_visualization}
\vspace{-0.4cm}
\end{figure}

To further demonstrate our model can effectively capture related information about the actor and action from the textual query, we conduct an additional experiment by randomly setting the word attention weights for the textual query during the model testing process. The experimental results are shown in Table \ref{table:random_weights}. It can be seen that the segmentation performance deteriorates notablely by assigning random attention weights for the textual query. This indicates that the language attention learning enables to capture key-word information for the actor matching and action matching in AAMN. In Fig. \ref{fig:random_visualization} (b) and (c), we present two testing examples with predicted attention weights and random attention weights, respectively. It can be observed that the predicted attention weights focus on key words about the actor and action, and thus the model can correctly localize and segment the referred actor. However, the model with random attention weights fails to localize the referred actor, as the model ignores meaningful information about the actor and action from the textual query.

\subsection{Comparison with State-of-the-art Methods}
\subsubsection{Single-Frame Segmentation from a Textual Query}

Since sparse annotations on A2D dataset, previous approaches \cite{gavrilyuk2018actor,wang2019asymmetric,wang2020context,mcintosh2018multi} take a video clip around the annotated frame and a textual query as inputs. They utilized pre-trained 3D CNN to extract the clip features, and conduct temporal average pooling to obtain a 2D feature map. The segmentation performance is evaluated on the annotated frame by upsampling the fused multi-modal features from the low resolution to high resolution by introducing 2D deconvolutional layers. While other approaches \cite{hu2016segmentation,li2017tracking} are proposed for image segmentation from a textual query, they directly take the annotated frame and a textual query as inputs and ignore the temporal modeling. Therefore, aforementioned approaches are single-frame segmentation models as they only output single-frame predictions at a time. Different from them, AAMN is a multi-frame model which outputs multi-frame predictions at a time. Specifically, by taking a video clip and a textual query as inputs, AAMN first localizes the tube that contains the referred actor and action in the video clip, and then feeds the tube into a FCN. The FCN is only trained with the annotated frame, \emph{i.e.}, middle frame, of the selected tube during the model learning. The FCN can predict all masks of the referred actor along the selected tube during the model testing. In this set of experiments, we present the segmentation results of AAMN on annotated frames as previous works.

% CMSA+
% a cross-modal self-attention module effectively captures the long-range dependencies between linguistic and visual contexts to produce a robust multi-modal feature representation for the referring segmentation, and cross-frame self-attention module  for extracting effective feature information from a clip of the video for the current frame.

% table 1: SOTA on A2D dataset
\begin{table*}[!htb]
%\small
\begin{threeparttable}
%\begin{center}
%\fontsize{6.5}{8}\selectfont
\caption{Comparison with the state-of-the-art methods on A2D Sentences}
\label{table:A2D_comparison}
\centering
\setlength{\tabcolsep}{3.1mm}{
\begin{tabular}{lccccccccccc}

\toprule
\multirow{2}{*}{\textbf{Methods}}&\multirow{2}{*}{Input}&\multicolumn{5}{c}{\textbf{Overlap}}&\textbf{mAP}&\multicolumn{2}{c}{\textbf{IoU}}&\multirow{2}{*}{Speed}\\
\cmidrule(lr){3-7}
\cmidrule(lr){8-8}
\cmidrule(lr){9-10}
& & P@0.5 & P@0.6 & P@0.7 & P@0.8 & P@0.9 & 0.5:0.95 & Overall & Mean &\\
\midrule
%Hu \emph{et al.}\cite{hu2016segmentation}        &RGB            & 7.7  & 3.9   & 0.8  & 0.0  & 0.0   & 2.0   & 21.3 & 12.8 \\
%Li \emph{et al.}\cite{li2017tracking}            &RGB                 & 10.8 & 6.2   & 2.0  & 0.3  & 0.0   & 3.3   & 24.8 & 14.4 \\
Hu \emph{et al.} \cite{hu2016segmentation}  &RGB            & 34.8 & 23.6  & 13.3 & 3.3  & 0.1   & 13.2  & 47.4 & 35.0  & 60.6 ms\\
Li \emph{et al.} \cite{li2017tracking}      &RGB            & 38.7 & 29.0  & 17.5 & 6.6  & 0.1   & 16.3  & 51.5 & 35.4  & - \\
Gavrilyuk \emph{et al.} \cite{gavrilyuk2018actor}    &RGB   & 47.5 & 34.7  & 21.1 & 8.0  & 0.2   & 19.8  & 53.6 & 42.1  & 95.6 ms \\
ACGA            \cite{wang2019asymmetric}  &RGB                    & 55.7 & 45.9  & 31.9 & 16.0 & 2.0   & 27.4  & 60.1 & 49.0  & 108.9 ms \\

CMDy            \cite{wang2020context}  &RGB        & 60.7 & 52.5  & 40.5 & 23.5 & 4.5  & 33.3 & 62.3 &53.1  & 125.3 ms \\

VT-Capsule      \cite{mcintosh2018multi}      &RGB                    & 52.6 & 45.0  & 34.5 & 20.7 & 3.6   & 30.3  & 56.8 & 46.0  & - \\

%Ye \emph{et al.} (TPAMI 2021) \cite{cmsa_pami}                  & RGB            & 46.7 & 38.5 & 27.9 & 13.6  & 1.7 & -  & 59.2  & 40.5 \\

AAMN (Ours)                    &RGB        &65.2  &58.0  &44.5  &20.8  &1.6  &34.8  &58.8 &51.3  & 446.6 ms \\
\cmidrule{1-11}

Gavrilyuk \emph{et al.} \cite{gavrilyuk2018actor}  &RGB+Flow    & 50.0 & 37.6  & 23.1 & 9.4  & 0.4   & 21.5  & 55.1 & 42.6  & 174.7 ms \\
ACGA                    \cite{wang2019asymmetric}  &RGB+Flow    & -    & -     & -    & -    & -     & 28.7  & 60.6 & 50.3  & 184.8 ms \\
CMDy                    \cite{wang2020context}     &RGB+Flow    & 63.3 & 55.0  & 42.8 & 25.7 &\textbf{4.9} & 34.2  & 62.9  &53.8  & 201.3 ms \\
AAMN (Ours)                       &RGB+Flow    &\textbf{68.9}	&\textbf{64.2}	&\textbf{54.5}	&\textbf{32.4}	&3.4	&\textbf{41.2}	 &\textbf{63.4}	&\textbf{55.6}  & 546.5 ms \\

\bottomrule
\end{tabular}}
%\end{center}
\end{threeparttable}
%\vspace{-0.1cm}
\end{table*}

% table 2: SOTA on JHMDB dataset
\begin{table*}[!htb]
\begin{threeparttable}
%\begin{center}
%\fontsize{6.5}{8}\selectfont
\caption{Comparison with the state-of-the-art methods on J-HMDB Sentences}
\label{table:J-HMDB_comparison}
\centering
\setlength{\tabcolsep}{4.0mm}{
\begin{tabular}{lccccccccc}

\toprule
\multirow{2}{*}{\textbf{Methods}} &\multirow{2}{*}{Input} &\multicolumn{5}{c}{\textbf{Overlap}}&\textbf{mAP}&\multicolumn{2}{c}{\textbf{IoU}}\\
\cmidrule(lr){3-7}
\cmidrule(lr){8-8}
\cmidrule(lr){9-10}
& & P@0.5 & P@0.6 & P@0.7 & P@0.8 & P@0.9 & 0.5:0.95 & Overall & Mean \\
\midrule
Hu \emph{et al.}  \cite{hu2016segmentation}    &RGB      & 63.3   & 35.0  & 8.5   & 0.2    & 0.0   & 17.8   & 54.6   & 52.8 \\
Li \emph{et al.}  \cite{li2017tracking}        &RGB      & 57.8   & 33.5  & 10.3  & 0.6    & 0.0   & 17.3   & 52.9   & 49.1 \\

%CMSA+CFSA \cite{cmsa_pami}                     & RGB     & 76.4   & 62.5  & 38.9  & 9.0  & 0.1  & -  & 62.8    & 58.1  \\

ACGA              \cite{wang2019asymmetric}        &RGB          & 75.6   & 56.4  & 28.7  & 3.4    & 0.0   & 28.9   & 57.6   & 58.4 \\

CMDy              \cite{wang2020context}           &RGB          & 74.2 & 58.7  & 31.6  & 4.7    & 0.0   & 30.1   & 55.4   & 57.6 \\

VT-Capsule \cite{mcintosh2018multi}     &RGB           & 67.7   & 51.3  & 28.3  & \textbf{5.1}   & 0.0    & 26.1   & 53.5    & 55.0 \\

%Ye \emph{et al.} (TPAMI 2021) \cite{cmsa_pami}            &RGB      & 76.0  & 60.4  & \textbf{36.4}  & \textbf{6.4}  & 0.0  & - & \textbf{60.6}  & 57.0 \\

AAMN (Ours)                    &RGB           & 74.6 & 57.3 & 25.6 & 1.5  & 0.0 & 27.8 & 55.6 & 56.3 \\
\cmidrule{1-10}
Gavrilyuk \emph{et al.}  \cite{gavrilyuk2018actor}   &RGB+Flow      & 69.9   & 46.0  & 17.3  & 1.4    & 0.0   & 23.3   & 54.1   & 54.2 \\
ACGA                     \cite{wang2019asymmetric}         &RGB+Flow     & -      & -     & -     & -      & -     & 29.5   & 57.9   & \textbf{59.1} \\
CMDy               \cite{wang2020context}            &RGB+Flow     & 76.6 & 60.0 & 32.9 & 5.0 & 0.0 & 30.8 & 56.2 & 58.0 \\

AAMN (Ours)                    &RGB+Flow      & \textbf{78.5} &\textbf{63.4} & \textbf{34.6}  & 3.8 & 0.0 & \textbf{32.0} & \textbf{59.3} & 59.0 \\
\bottomrule
\end{tabular}}
%\end{center}
\end{threeparttable}
\vspace{-0.2cm}
\end{table*}

In Table \ref{table:A2D_comparison}, we present comparisons of our model with previous methods on A2D Sentences. For fair comparison with methods that only take the RGB clip as input, we remove the Flow stream and the gated fusion in the action module. The exprimental results are shown in Line 7 of Table \ref{table:A2D_comparison}. It can be observed that our method outperforms previous methods under most metrics, especially for the \emph{Overlap} in \emph{P@0.5}, \emph{P@0.6}, and \emph{P@0.7}, which surpass the best results by $4.5\%$, $5.5\%$, and $4.0\%$, respectively. Previous works \cite{fusionseg,monet,2019Lucid} have demonstrated that integrating the optical flow is beneficial to improve the performance of video object segmentation. The methods \cite{gavrilyuk2018actor,wang2019asymmetric} also explored two-stream models that take RGB and Flow clips as inputs. Their methods fused RGB and Flow streams by computing a weighted average of response maps from each stream. We also extend the single-stream model in \cite{wang2020context} to a two-stream model by adopting the same way to fuse RGB and Flow streams. Our method is a two stream model that takes RGB and Flow clips as inputs. The optical flow is introduced to localize the action which performs by the referred actor. The experimental results of our full model are shown in Line 11 of Table \ref{table:A2D_comparison}. We can see that the performance of our model significantly outperforms state-of-the-art results from single-stream models, and our model achieves the best performances compared with previous two-stream models. Specifically, compared with the best results from two-stream model in \cite{wang2020context}, our model brings $5.6\%$, $9.2\%$, $11.7\%$ and $6.7\%$ improvements under metrics of \emph{P@0.5}, \emph{P@0.6}, \emph{P@0.7}, and \emph{P@0.8}. The performance under metrics of \emph{mAP}, \emph{Overall IoU}, and \emph{Mean IoU} also achieves $7.0\%$, $0.5\%$, and $1.8\%$ improvements, respectively.

We note that the performance of our model is slightly worse than the best results for the \emph{Overlap} in \emph{P@0.9}, and the performance improvement is small in \emph{Overall IoU}. The main reason is that we adopt proposal-based approach \cite{he2017mask} to predict the mask of actor on a $28\times{28}$ grid irrespective of actor size. Such a grid is sufficient for small actors, but for large actors further upsampling with bilinear interpolation will over-smooth the fine-grained details \cite{pointRend}. While previous works \cite{gavrilyuk2018actor,wang2019asymmetric,wang2020context,mcintosh2018multi} predict the mask of the actor by gradually upsampling the full feature map with multiple deconlutional layers. They introduce skip-connections in the model \cite{mcintosh2018multi} or supervisions on different resolutions during model training \cite{gavrilyuk2018actor,wang2019asymmetric}. Therefore, these methods tend to preserve fine-grained details and possess higher overlaps for both small and large actors.

In addition, we also compare the inference speed for different methods in Table \ref{table:A2D_comparison}. The inference speed indicates the total time for the method predicts the final segmentation masks from a clip-query pair. It can be observed that the inference speed of our method is slower than other methods. This is because our method requires more time for proposal detection per frame in the input video clip. However, our method provides a new solution for text-based video segmentation from another perspective, which is significantly different from existing methods. And our method is more efficient than other methods, as they only predict one-frame segmentation results from the clip-query pair, while our method can segment all frames of the video clip.

Following \cite{gavrilyuk2018actor}, we also evaluate our method on J-HMDB Sentences to validate the generalization ability of our method. In our experiments, the testing model is pre-trained on A2D Sentences, and J-HMDB Sentences is used as a testing dataset. For fair comparisons, we also uniformly sample three clips for each video as previous works. The experimental results are shown in Table \ref{table:J-HMDB_comparison}. In Line 6 of Table \ref{table:J-HMDB_comparison}, we show experimental results that we remove the Flow stream in our model. It can be observed that our method achieves superior results compared with previous methods. In Line 10 of Table \ref{table:J-HMDB_comparison}, we show the experimental results from full model which takes RGB and Flow clips as inputs. Compared with two-stream models, our method achieves state-of-the-art performance under most metrics. Specifically, compared with the best results from two-stream method in \cite{wang2020context}, our method achieves  $1.9\%$, $3.4\%$ and $1.7\%$ improvements in terms of \emph{P@0.5}, \emph{P@0.6} and \emph{P@0.7}. The \emph{mAP} and \emph{Overall IoU} also surpass the best results by $1.2\%$ and $1.4\%$, respectively. The results in terms of \emph{P@0.8} and \emph{Mean IoU} are slightly worse than the best results. It is probably caused by the backbone that cannot extract fine representations without finetuning on J-HMDB Sentences.

To validate the importance of temporal modeling by associating objects cross frames for text-based video segmentation, we also compare our method with some state-of-the-art referring expression segmentation methods on A2D Sentences. The experimental results are shown in Table \ref{table:referseg}. It can be observed that our method achieves superior performance than these methods. This demonstrates that the temporal modeling by associating objects cross frames is truly helpful to localize and segment the referred actor in the video.

\subsubsection{Full Segmentation from a Textual Query}

Previous methods \cite{gavrilyuk2018actor,wang2019asymmetric,wang2020context} are very time consuming for full video segmentation as they only predict one frame segmentation at a time. McIntosh \emph{et al.} \cite{mcintosh2018multi} further replaced 2D deconvolutional layers with 3D deconvolutional layers in their decoder network, which enables the model to predict multi-frame segmentation results at a time. Similar to \cite{mcintosh2018multi}, our method is a multi-frame segmentation model, which can be applied to full video segmentation by segmenting video clip-by-clip.

% table : Ñ¡²»Í¬µÄÖ¡ £¨2L+1£©
\begin{table}[!tb]
%\small
\begin{threeparttable}
%\begin{center}
%\fontsize{6.5}{8}\selectfont
\caption{Comparison with state-of-the-art referring expression segmentation methods}
\label{table:referseg}
\centering
\setlength{\tabcolsep}{0.58mm}{
\begin{tabular}{lcccccccc}

\toprule
\multirow{2}{*}{\textbf{Models}}&\multicolumn{5}{c}{\textbf{Overlap}}&\textbf{mAP}&\multicolumn{2}{c}{\textbf{IoU}}\\
\cmidrule(lr){2-6}
\cmidrule(lr){7-7}
\cmidrule(lr){8-9}
& P@0.5 & P@0.6 & P@0.7 & P@0.8 & P@0.9 & 0.5:0.95 & Overall & Mean \\
\midrule

RRN    \cite{li2018referring}   & 42.0 	 & 35.5	  &	27.5   & 15.4  & 2.5     & 22.4	   & 56.4	 & 37.3  \\
CMSA       \cite{cmsa_pami}     & 46.7   & 38.5   &	27.9   & 13.6  & 1.7     & -	   & 59.2	 & 40.5 \\
BRINet \cite{hu2020bi}          & 55.7 	 & 46.4	  &	33.0   & 17.5  & 3.3     & 28.3	   & 63.0	 & 47.4  \\
CMPC   \cite{liu2021cross}  & 59.0   & 52.7   & 43.4   & 28.4   & \textbf{6.8}    & 35.1    & \textbf{64.9}    & 51.5 \\
Ours   & \textbf{68.9}	& \textbf{64.2}	   & \textbf{54.5}	& \textbf{32.4}	 & 3.4	 & \textbf{41.2}	   & 63.4	 & \textbf{55.6} \\

\bottomrule
\end{tabular}}
%\end{center}
\end{threeparttable}
\vspace{-0.1cm}
\end{table}

% table 1: full video segmentation on A2D
\begin{table}[!tb]
%\small
\begin{threeparttable}
%\begin{center}
%\fontsize{6.5}{8}\selectfont
\caption{Experimental results on A2D Sentences for full video segmentation}
\label{table:a2d_full}
\centering
\setlength{\tabcolsep}{0.15mm}{
\begin{tabular}{lcccccccc}

\toprule
\multirow{2}{*}{\textbf{Methods}}&\multicolumn{5}{c}{\textbf{Overlap}}&\textbf{mAP}&\multicolumn{2}{c}{\textbf{IoU}}\\
\cmidrule(lr){2-6}
\cmidrule(lr){7-7}
\cmidrule(lr){8-9}
&  P@0.5 & P@0.6 & P@0.7 & P@0.8 & P@0.9 & 0.5:0.95 & Overall & Mean \\
\midrule

VT-Capsule  (pixel)                &9.6     &1.6    &\textbf{0.4}    &0.0   &0.0   &1.8     &34.4    &26.6 \\
AAMN (pixel)                                    &\textbf{13.9}     &\textbf{2.8}    &0.3    &0.0   &0.0   &\textbf{2.5}     &\textbf{35.6}    &\textbf{29.8} \\
\cmidrule{1-9}
VT-Capsule  (bbox)                 &41.9    &33.3   &22.2   &\textbf{10.0}  &0.1   &21.2    &51.5    &41.3 \\
AAMN (bbox)                                     &\textbf{46.7}    &\textbf{37.8}   &\textbf{24.0}   &9.1   &\textbf{0.2}   &\textbf{21.5}    &\textbf{52.1}    &\textbf{43.5} \\
%\cite{mcintosh2018multi} (All frames)    &45.6    &37.4   &25.3   &10.0  &0.4   &23.3    &55.7    &41.8 \\

\bottomrule
\end{tabular}}
%\end{center}
\end{threeparttable}
\vspace{-0.3cm}
\end{table}

To explore full video segmentation, McIntosh \emph{et al.} further annotated actors with bounding box in all frames of the video. They utilized bounding boxes to train and evaluate their model. The predictions of the model is block-like, which is not precise enough for pixel-level segmentation of the actor. Different from them, we only train our model with ground truth masks in key frames. For ease of comparison with \cite{mcintosh2018multi}, we also evaluate our model with bounding box annotations in all frames of the video. The experimental results are shown in Table \ref{table:a2d_full}. The performance is evaluated with pixel-wise segmentation output in the first and second lines. In the third and last lines, the bounding box is placed around the fine-grained segmentation masks for evaluation. As observed from Table \ref{table:a2d_full}, our method achieves superior performance compared with previous method in \cite{mcintosh2018multi} for the full video segmentation.

\vspace{-0.5cm}
\subsection{Visualization and Failure Cases}
\subsubsection{Qualitative Results}
\begin{figure}[!tb]
\centering
\includegraphics[width=\linewidth]{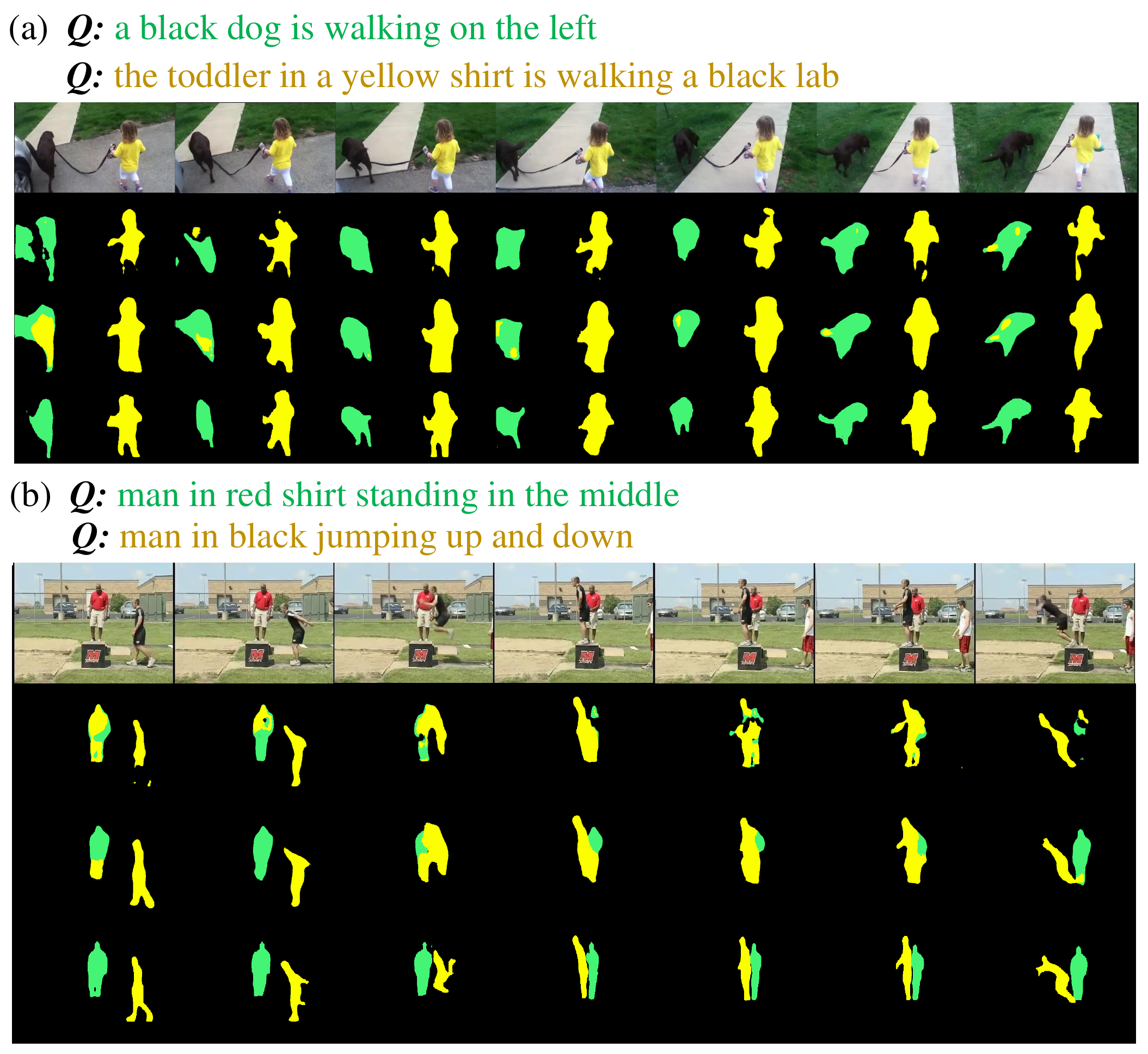}
\caption{The qualitative results shown in sampled frames for several videos from A2D Sentences (Best viewed in color). For each example, from the first row to the fourth row show the sampled video frames, the segmentation results from \cite{gavrilyuk2018actor}, the segmentation results from \cite{wang2019asymmetric}, and the segmentation results from our AAMN. The queries are given on the top and their colors correspond to mask colors.}
\label{fig:visualization}
\vspace{-0.4cm}
\end{figure}

To qualitatively validate the effectiveness of our method, we provide some segmentation results in sampled frames for two videos on A2D Sentences. The visualization results are shown in Fig. \ref{fig:visualization}, where mask colors correspond to query colors. For each example, the first row shows input video frames. The second row shows the results produced by the method in \cite{gavrilyuk2018actor}. The third row shows the results produced by the method in \cite{wang2019asymmetric}. The last row shows the results produced by our method. It can be observed that our method can produce better visual qualitative results than previous methods, and more details of the actor are presented in our results. For example, the boundaries of the toddler in the first video and two man in the second video are clearer than the results produced by two other methods. Besides, segmentation results from methods in \cite{gavrilyuk2018actor, wang2019asymmetric} involve confused pixels between different actors in two videos. Such confused pixels indicate that these methods cannot precisely distinguish the referred actors in sampled frames. Different from them, our method enables to correctly distinguish the referred actors and generate more precise segmentation masks.

\subsubsection{Failure Cases}
\begin{figure}[!tb]
\centering
\includegraphics[width=\linewidth]{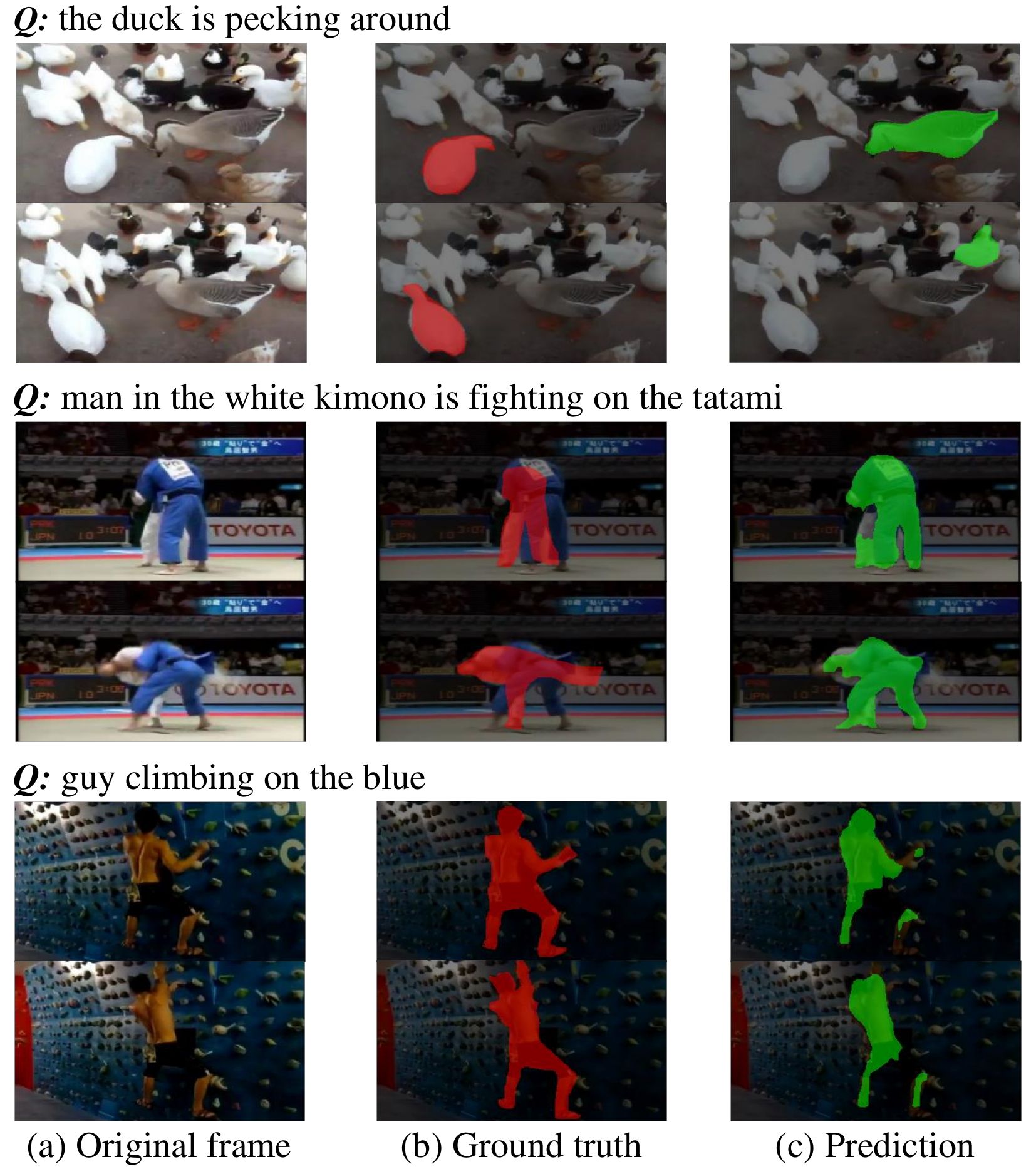}
\caption{Some failure cases on A2D Sentences. (a) The original video frame. (b) The ground truth. (c) The failures of prediction caused by the ambiguous description for the target actor (the first video), the occlusion between different instances (the second video), and the blurry boundary of the target actor (the third video). Note that we sample two annotated frames from each video.}
\label{fig:failure_cases}
\vspace{-0.4cm}
\end{figure}

Despite our AAMN has achieved significant improvements than other methods, there are still some challenging scenarios that cannot be handled well. Some typical failure cases from A2D Sentences are shown in Fig. \ref{fig:failure_cases}. We argue that these failure cases are mainly due to the following three aspects. \emph{First}, the ambiguous description for the target actor. For example, in the first video, it aims to segment the duck on the bottom-left with the textual query \emph{``the duck is pecking around''}. Our model suffers from difficulty to identify the referred duck and ducks around it, because they have similar appearance and motion information. \emph{Second}, the occlusion between different instances. The occlusion in the video makes it difficult to correctly localize the target actor with inaccurate detection results. It also deteriorates the segmentation results as the model cannot distinguish different instances within the localized region. For example, in the second video, our model fails to segment the man in white as he is partially occluded by the man in blue. \emph{Third}, the blurry boundary of the target actor. Our model fails to produce sharper predictions around actor boundaries when the actor involves similar colors with background or fast motion in the video. For example, in the third video, our predictions cannot precisely segment the right thigh of the man because the part of right thigh is hard to be distinguished from background. Some of failure cases caused by such blurry boundary may be improved by enhancing shape information from adjacent frames.

\section{Conclusion}
In this paper, we have studied a challenging problem which is text-based video segmentation. To deal with this problem, we proposed an actor and action modular network to achieve video-query symmetrical matching. We first generate actor-/action-related tubes for input video clip, and learn the actor-/action-related language representations from the textual query. Then we use the modular network to locate the target tube, which involves the referred actor and action. Finally, the target tube is fed into a tiny FCN to predict the binary masks of the actor within the tube. We conducted extensive experiments to demonstrate the effectiveness of our method, and our method achieves state-of-the-art performance on two benchmark datasets. To explore full video segmentation, McIntosh \emph{et al.} annotated actors with bounding boxes in all frames of the video. But the segmentation results are not precise enough for the model trained with bounding boxes. Considering pixel-level annotation for all frames of video is time-consuming and cost expensive, we will explore weakly supervised text-based video segmentation in the future work.

% if have a single appendix:
%\appendix[Proof of the Zonklar Equations]
% or
%\appendix  % for no appendix heading
% do not use \section anymore after \appendix, only \section*
% is possibly needed

% use appendices with more than one appendix
% then use \section to start each appendix
% you must declare a \section before using any
% \subsection or using \label (\appendices by itself
% starts a section numbered zero.)
%

%\appendices
%\section{Proof of the First Zonklar Equation}
%Appendix one text goes here.

% you can choose not to have a title for an appendix
% if you want by leaving the argument blank
%\section{}
%Appendix two text goes here.

% Can use something like this to put references on a page
% by themselves when using endfloat and the captionsoff option.
\ifCLASSOPTIONcaptionsoff
  \newpage
\fi

% trigger a \newpage just before the given reference
% number - used to balance the columns on the last page
% adjust value as needed - may need to be readjusted if
% the document is modified later
%\IEEEtriggeratref{8}
% The "triggered" command can be changed if desired:
%\IEEEtriggercmd{\enlargethispage{-5in}}

% references section

% can use a bibliography generated by BibTeX as a .bbl file
% BibTeX documentation can be easily obtained at:
% http://mirror.ctan.org/biblio/bibtex/contrib/doc/
% The IEEEtran BibTeX style support page is at:
% http://www.michaelshell.org/tex/ieeetran/bibtex/
\bibliographystyle{IEEEtran}
% argument is your BibTeX string definitions and bibliography database(s)
%\small
\bibliography{IEEEabrv,IEEEfull}

% <OR> manually copy in the resultant .bbl file
% set second argument of \begin to the number of references
% (used to reserve space for the reference number labels box)
%\begin{thebibliography}{1}
%\bibitem{IEEEhowto:kopka}
%H.~Kopka and P.~W. Daly, \emph{A Guide to \LaTeX}, 3rd~ed.\hskip 1em plus
%  0.5em minus 0.4em\relax Harlow, England: Addison-Wesley, 1999.
%\end{thebibliography}

% biography section
%
% If you have an EPS/PDF photo (graphicx package needed) extra braces are
% needed around the contents of the optional argument to biography to prevent
% the LaTeX parser from getting confused when it sees the complicated
% \includegraphics command within an optional argument. (You could create
% your own custom macro containing the \includegraphics command to make things
% simpler here.)
%\begin{IEEEbiography}[{\includegraphics[width=1in,height=1.25in,clip,keepaspectratio]{mshell}}]{Michael Shell}
% or if you just want to reserve a space for a photo:

% if you will not have a photo at all:
%\begin{IEEEbiographynophoto}{John Doe}
%Biography text here.
%\end{IEEEbiographynophoto}

% insert where needed to balance the two columns on the last page with
% biographies
%\newpage

% You can push biographies down or up by placing
% a \vfill before or after them. The appropriate
% use of \vfill depends on what kind of text is
% on the last page and whether or not the columns
% are being equalized.

%\vfill

% Can be used to pull up biographies so that the bottom of the last one
% is flush with the other column.
%\enlargethispage{-5in}

% that's all folks
\end{document}